\documentclass{article}

\usepackage{arxiv}

\usepackage{lineno}

\usepackage[utf8]{inputenc} 
\usepackage[T1]{fontenc}    
\usepackage{hyperref}       
\usepackage{url}            
\usepackage{booktabs}       
\usepackage{amsfonts}       
\usepackage{nicefrac}       
\usepackage{microtype}      
\usepackage{lipsum}
\usepackage{graphicx}
\usepackage{authblk}

\usepackage{amsmath}
\usepackage{multirow}

\usepackage{subcaption}
\usepackage[ruled,linesnumbered,algo2e]{algorithm2e}
\usepackage[flushleft]{threeparttable}
\usepackage{appendix}
\usepackage{supertabular}

\title{RCT: Resource Constrained Training for Edge AI}

\author[1]{Tian Huang}
\author[1]{Tao Luo}
\author[1]{Ming Yan}
\author[1]{Joey Tianyi Zhou}
\author[1]{Rick Goh}
\affil[1]{Institute of High Performance Computing, A*STAR}
\affil[]{\{huang\_tian, luo\_tao, yan\_ming, joey\_zhou, gohsm\}@ihpc.a-star.edu.sg}

\begin{document}

\maketitle

\begin{abstract}
Neural networks training on edge terminals is essential for edge AI computing, which needs to be adaptive to evolving environment. Quantised models can efficiently run on edge devices, but existing training methods for these compact models are designed to run on powerful servers with abundant memory and energy budget. For example, quantisation-aware training (QAT) method involves two copies of model parameters, which is usually beyond the capacity of on-chip memory in edge devices. Data movement between off-chip and on-chip memory is energy demanding as well. The resource requirements are trivial for powerful servers, but critical for edge devices. To mitigate these issues, We propose Resource Constrained Training (RCT). RCT only keeps a quantised model throughout the training, so that the memory requirements for model parameters in training is reduced. It adjusts per-layer bitwidth dynamically in order to save energy when a model can learn effectively with lower precision. We carry out experiments with representative models and tasks in image application and natural language processing. 
Experiments show that RCT saves more than 86\% energy for General Matrix Multiply (GEMM) and saves more than 46\% memory for model parameters, with limited accuracy loss. Comparing with QAT-based method, RCT saves about half of energy on moving model parameters.
\end{abstract}

\section{Introduction}

Edge computing emerges as an attractive alternative to cloud computing, for its advantages in data privacy, response latency and energy saving for data transmissions \cite{chen2019review}. Deep learning-based intelligence, as another growing trend, have started to change many aspects of people’s lives. Well trained deep learning models are able to achieve record-breaking predictive performance \cite{krizhevsky2012imagenet, hinton2012deep, silver2016mastering}, but we have witnessed increasing demand for the model to learn in-situ, for the purpose of personalisation or adaptation to evolving environment. Comparing to cloud-based deep learning framework, training a neural network on edge device reflects the very idea of edge computing, and benefits from energy savings in data transmission, low response latency and enhanced privacy.


Quantisation is a popular technique that improves energy efficiency and speed of a neural network. There are many free and commercialised tools for converting well-trained neural network to a quantised one. But training a quantised model is tricky. Quantisation-aware training (QAT) \cite{krishnamoorthi2018quantizing} is a big family of methods that matches the requirement of in-situ learning. Most of QAT-based training methods are based on the assumption that resources like memory and energy are abundant.


\begin{figure}[htb]
     \centering
     \begin{subfigure}[b]{0.49\linewidth}
         \centering
         \includegraphics[width=\linewidth]{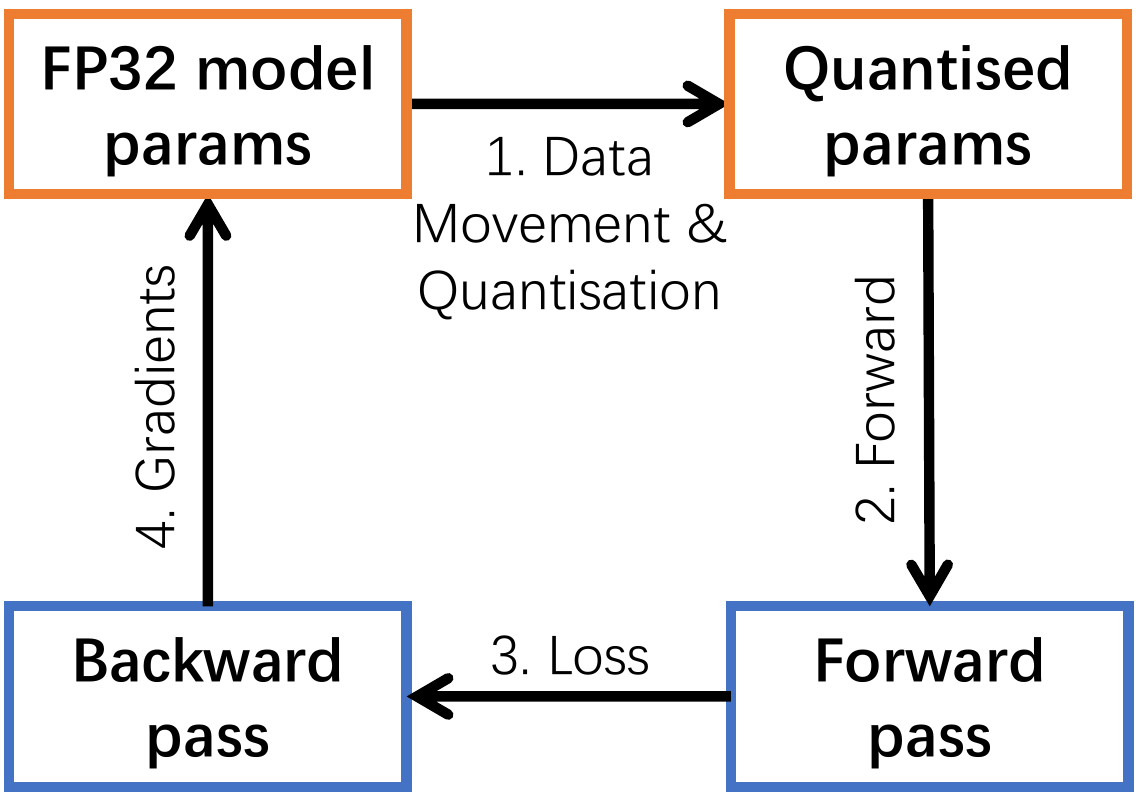}
         \caption{QAT-based methods}
         \label{fig:qat_diagram}
     \end{subfigure}
     \hfill
     \begin{subfigure}[b]{0.49\linewidth}
         \centering
         \includegraphics[width=\linewidth]{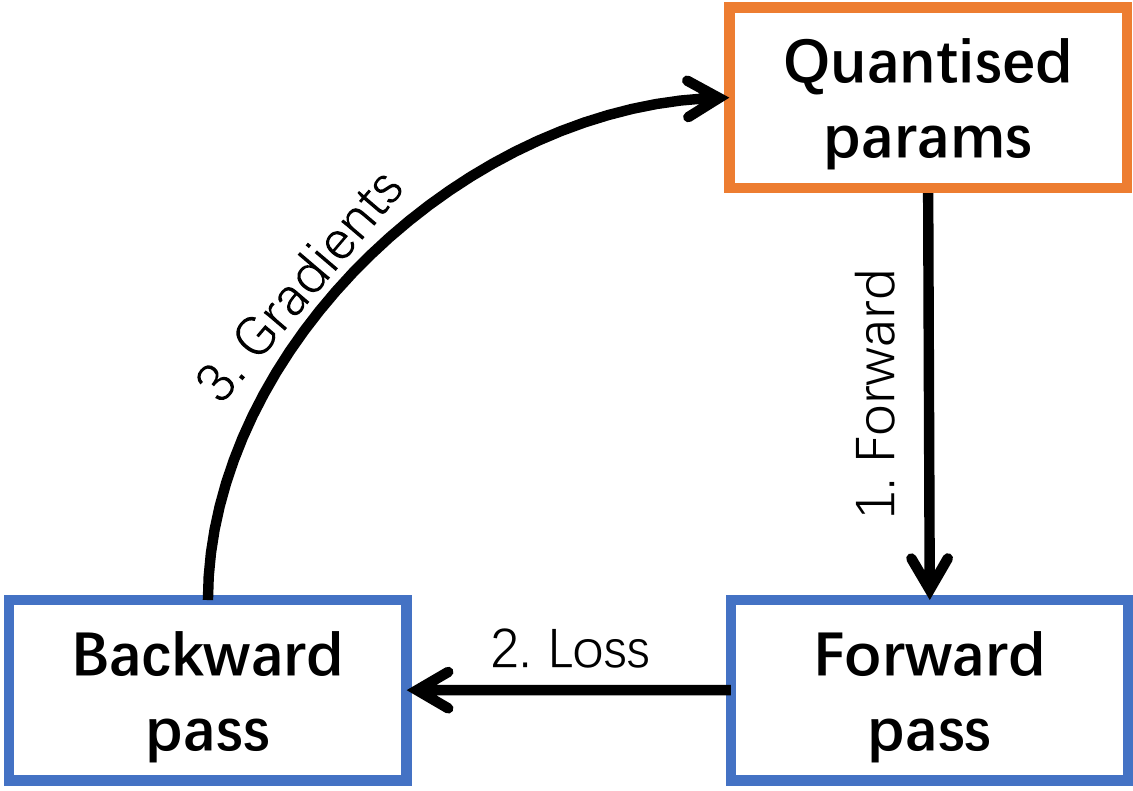}
         \caption{Our method}
         \label{fig:rct_diagram}
     \end{subfigure}

        \caption{Comparison between QAT-based methods and RCT. (a) QAT-based methods involve two copies of model parameters. (b) Our method RCT only keeps a quantised model.}
        \label{fig:training_diagram_comprison}
\end{figure}

QAT-based methods involve two copies of model parameters during training, as shown in Figure \ref{fig:qat_diagram}. (1) At first stage, floating-point model travels from memory to processor for quantisation. (2),(3) Then quantised model takes part in inference and generate loss. (4) At the stage of model update, the floating-point model has to travel from memory to processor again such that the gradients can be added to the floating point parameters. The update is usually followed by another write-back to the memory. Such training method is not friendly for edge AI from two aspects. First, on-chip memory of edge devices (e.g. cache or buffer in a processor) is usually limited in size such that the floating-point model parameters have to stay in off-chip memory (e.g. DRAM), which is much slower than on-chip memory. Second, data movement is energy demanding. Battery-powered edge devices may not have enough energy to support two copies of a model and frequent data movement between off-chip memory and processors.

As shown in Figure \ref{fig:rct_diagram}, if we only keep a quantised model, we saves the memory for floating-point parameters. Since quantised model is usually compact enough to stay in on-chip memory throughout training, we can also save energy for related data movement. 

The challenge is that with quantised parameters alone, a model does not always learn effectively. A model with low-precision parameters may learn without difficulty at the beginning of the training process. As training loss and gradients decrease along with the training progress, quantisation underflow happens more often in the model update step, and slows down or even put a stall to the training progress. Allocating more bitwidth could mitigate the underflow issue and sustain the overall training efficiency.


Based on this observation, we propose a heuristic method called \textbf{Resource Constrained Training (RCT)}, which applies a per-layer metrics that reflects how often quantisation underflow is happening. By evaluating the metrics, RCT understands how effective a layer learns with given precision. RCT allocates more bitwidth to the parameters of a layer if underflow happens very often. The per-layer bitwidth of a model changes dynamically throughout the training process. Below shows a list of our contributions:

\begin{itemize}
  \item We propose a heuristic training method RCT, which is aware of the training progress and adjusts per-layer bitwidth dynamically to save energy for training.
  \item RCT only keeps quantised parameters. In this way, we reduce the memory requirements for model parameters in training.
  \item RCT provides an application specific hyper-parameter that achieves trade-off between training energy, memory usage, and accuracy.
\end{itemize}

RCT finds the per-layer bitwidth configuration that helps the model learn effectively with quantised parameteres only. It benefits edge devices that have to learn in-situ frequently after deployment. In the experiment we apply RCT to representative CNN models on image classification tasks. 
We also evaluate RCT with representative NLP applications. 
Comparing to non-quantisation baseline, RCT saves more than 86\% energy for General Matrix Multiply (GEMM) and saves more than 46\% memory for model parameters, with limited accuracy loss. 
Comparing with other QAT-based methods, RCT saves more than half of memory for model parameters and related energy for transmitting model parameters.


\section{Related Work}

There are two kinds of popular quantisation methods, Quantisation-aware training (QAT) and post-training quantisation \cite{krishnamoorthi2018quantizing}. QAT emulates inference-time quantisation during training, which helps in maintaining accuracy as compared to 16-bit or 32-bit floating-point baselines. The recent progress in QAT has allowed the realization of results comparable to a baseline for as low as 2–4 bits per parameter \cite{shkolnik2020robust,gong2019differentiable,jin2019towards,yang2019quantization,baskin2018nice,zhang2018lq,li2016ternary} and show decent performance even for single bit (binary) parameters \cite{bai2020binarybert,kim2020binaryduo,liu2019rbcn,peng2019bdnn,rastegari2016xnor,zhou2016dorefa}. There are QAT-based methods that efficiently exploit various processing units \cite{micikevicius2017mixed,banner2018scalable,wu2018training,yang2020training,zafrir2019q8bert,kim2021bert}, as well as the ones that allocate per-layer bitwidth \cite{wu2018mixed,wang2019haq,shen2020q,zhou2018adaptive} to improve the performance of a quantised model. Apart from image related applications, QAT-based methods has also been applied to Natural Language Processing (NLP). \cite{zafrir2019q8bert,shen2020q,bai2020binarybert,kim2021bert}

QAT-based methods can train a quantised model. 
However, these QAT-based methods involves two copies of model parameters, as they are based on the assumption that the machine for training has abundant memory and electricity supply. This is usually not true for edge devices. We propose RCT, which only has one copy of model parameters. It saves memory, as well as the energy for related data movement.

The post-training quantisation \cite{gong2018highly,mellempudi2017ternary,choukroun2019low,nayak2019bit,fang2020near,zadeh2020gobo} converts a well-trained model into quantised one. By definition, post-training quantisation methods convert models but do not train them. It does not meet the requirements of in-situ learning, in which quantised models continue to evolve on edge devices.

Imposing \emph{sparsity} in neural network is an effective way of compression \cite{han2015deep, frankle2018lottery, dettmers2019sparse, gale2019state, liu2018rethinking}. In general, sparsity comes in the cost of irregular memory access patterns and unbalanced computation workload, \cite{guo2017survey,sohoni2019low} which raise resource utilisation issues on a resource constrained devices. Other techniques, e.g. \emph{Knowledge Distillation} \cite{gou2020knowledge}, \emph{Low-rank Factorisation} \cite{sainath2013low,li2018constrained}, \emph{Fast Convolution} \cite{zhang2017frequency}, save memory and/or energy significantly for inference. These methods only conduct compression but do not train the model after it is compressed.

\section{Resource Constrained Training}

In this section we describe the details of RCT. We explain why QAT-based methods involve two copies of model parameters, and how RCT can train effectively with only one quantised model. 

\subsection{Motivation}

Low precision integer arithmetic is more efficient than a floating point arithmetic in terms of energy and memory, but quantisation magnifies the underflow issue. \cite{krishnamoorthi2018quantizing} Equation \ref{eq:weight_update} shows the update process of a weight in fp32 format:

\begin{equation}
\label{eq:weight_update}
w_{ij} := w_{ij} - lr * g_{ij}
\end{equation}

$w_{ij}$ is the $j$-th weight of $i$-th layer, $lr$ is the learning rate, $g_{ij}$ is the corresponding gradient of the $w_{ij}$. Ideally, at each training step, a weight changes by $lr * g_{ij}$. As we apply $k$-bit quantisation to a tensor of weights, too small changes cannot be represented by the weight. We refer to this minimum resolution as $\epsilon$, which is formalised as follow:

\begin{equation}
\label{eq:epsilon}
\epsilon _i= \frac{max(W_i)-min(W_i)}{2^k-1}
\end{equation}

$W_i$ is a tensor of weights of $i$-th layer. $k$ is the precision, or the bitwidth for the tensor. The updates of a weight $w_{ij}$ is regulated by $\epsilon _i$, which can be formalised as the following equation:

\begin{equation}
w_{ij} := w_{ij} - \left \lfloor \frac{lr * g_{ij}}{\epsilon _i} \right \rfloor * \epsilon _i
\label{eq:weight_update_condition}
\end{equation}

As a result, $lr * g_{ij}$ is quantised to discrete states, which is equivalent to that of $w_{ij}$. $lr * g_{ij}$ has to be at least larger than $\epsilon _i$, otherwise quantisation underflow happens. For high precision training (e.g. 32-bit) $\epsilon$ is small, $lr * g_{ij}$ is larger than $\epsilon _i$ for most of cases. As the precision decreases, $\epsilon _i$ increases, quantisation underflow happens more often, putting higher resistance on weights updating. 

Gradient is the horse power that drives the training process forward. Low-precision training put restrictions on the horse power and slow down the training process. We have the observation that in low precision network, gradients vanishes quicker than that of a network with higher precision. This is because quantisation underflow happens more often in low precision model and guides the model into local minima. Some of layers have larger $\epsilon$ and are harder to be updated than others. As the training loss decreases, These layers with larger $\epsilon$ suffer more quantisation underflow than before, stepping onto a slippery slope. As the training process moves forward, quantisation underflow freezes more parameters, driving the training into a dead state.

Through the observation, we understand that mitigating the negative effect of quantisation underflow is critical to the training progress. Next we aim to find out how often quantisation underflow happens to a quantised layer.

\subsection{Metrics for Underflow}

QAT-based methods keep an fp32 model in order to avoid underflow, in the cost of memory usage, energy for data movement and additional training iterations. We aim to find low precision configurations that save memory and energy at the same time. The key is to understand how often quantisation underflow happens to a layer, We propose a metric called $Gavg$, which is formalised as Equation \ref{eq:gavg_metric}.

\begin{equation}
\label{eq:gavg_metric}
Gavg_i = \frac{1}{N_i}\sum_{j=0}^{N_i} \left | \frac{g_{ij}}{\epsilon _i} \right |
\end{equation}

$N_i$ represents the number of parameters in a tensor, $Gavg_i$ indicates how large a gradient is related to the minimum resolution of parameters in $i$-th layer. The larger the $Gavg$ is, the less likely the parameters remains unchanged during model update. If $Gavg$ approaches zero, that means the layer suffers serious quantisation underflow problem and does not update for most of times. The bitwidth is the key to prevent $Gavg$ from approaching zero. A higher bitwidth leads to lower $\epsilon$ and higher $Gavg$, which means the parameters are easier to update.  

Although we use weights to describe the concept of quantisation underflow and metrics, $Gavg$ applies to other parameters that need to be learned during training, e.g. bias, the clipping point of activation and gradient. In the metric $Gavg$, we do not include other factors like learning rate or momentum in the metric so that user can use other training tricks or sophisticated optimisers on top of our training method.

\subsection{Bitwidth Adjustment Policy}

The metric $Gavg$ is a measurement for balancing requirements between training energy, memory usage, and accuracy. It is well known that bitwidth is directly related to the energy, memory and accuracy. Using fixed bitwidth across the whole network may not meets the requirement of each layer. On comparison, a single threshold for $Gavg$ is able to produce different bitwidth configuration according to the distribution of parameters and their gradients.

We propose a bitwidth adjustment policy based on the metric $Gavg$. We start with a naive policy, which is to make sure that $Gavg$ of all layers are within a pre-defined range. A C-styled description of the policy is presented below.

\begin{algorithm2e}[h]
    \SetAlgoNoLine
    \SetKwFor{For}{for (}{) $\lbrace$}{$\rbrace$}
    \SetKwFor{If}{if (}{) $\lbrace$}{$\rbrace$}
    \SetKwInOut{Input}{input}
    \SetKwInOut{Output}{output}
    \Input{$k_{0\ldots M-1}$, $Gavg_{0\ldots M-1}$, $T_{min}$, $T_{max}$, }
    \Output{$k_{0\ldots M-1}$}
    \For {$i = 0;\ i < M;\ i = i + 1$}{
        \If{$Gavg_i < T_{min}$ \&\& $k_i < 32$}{
            $k_i := k_i + 1$\;
        }
        \If{$Gavg_i > T_{max}$ \&\& $k_i > 2$}{
            $k_i := k_i - 1$\;
        }
    }
    \Return{$k_{0\ldots M-1}$}\;
    \caption{Bitwidth Adjustment Policy}
    \label{alg:policy}
\end{algorithm2e}

In Algorithm \ref{alg:policy},  $k_{0\ldots M-1}$ and $Gavg_{0\ldots M-1}$ represents the bitwidth and metrics of all $M$ layers. $T_{min}$, $T_{max}$ are the upper and lower limit. Algorithm \ref{alg:policy} increase the bitwidth of a layer when its $Gavg < T_{min}$ and decrease the bitwidth when $Gavg > T_{max}$. The lower limit ensures all layers learn effective, whereas the upper limit is for saving energy cost and memory usage on those parameters that are very easy to update (e.g. due to its small range or large gradients).

\begin{algorithm2e}[h]
    \SetAlgoNoLine
    \SetKwFor{For}{for (}{) $\lbrace$}{$\rbrace$}
    \SetKwFor{If}{if (}{) $\lbrace$}{$\rbrace$}
    \SetKwInOut{Input}{input}
    \SetKwInOut{Output}{output}
    Initialise the model with low bitwidth, e.g. $k=8$ \; 
    \For {$epoch = 0;\ epoch < 200;\ epoch++$}{
        \For {$i = 0;\ i < len(train\_loader);\ i++$}{
            Forward propagation \;
            Loss back propagation \;
            \If{$i \% \textrm{INTERVAL} == 0$}{
                Evaluate $Gavg$ using Equation \ref{eq:gavg_metric} \;
                Adjust bitwidth using Algorithm \ref{alg:policy} \;
            }
            Update model parameters \;
        }
    }
    \caption{Training with RCT}
    \label{alg:training}
\end{algorithm2e}

A typical workflow of RCT is described in Algorithm \ref{alg:training}. A training starts with a low-precision model. The metrics evaluation and bitwidth adjustment happen between back propagation and model update. The evaluation and adjustment do not have to be done for each training iteration. A few times, say 10, in each epoch suffice to help the quantised training catch up with the progress of fp32 training. 

We emphasise that the model parameters in RCT are quantised. QAT-based methods quantise the fp32 parameters in Forward propagation in Line 4 of Algorithm \ref{alg:training}. In RCT, we only keep quantised parameters, so there is no parameter quantisation in forward pass at all.

\begin{figure}
\centering
\includegraphics[width=0.6\linewidth]{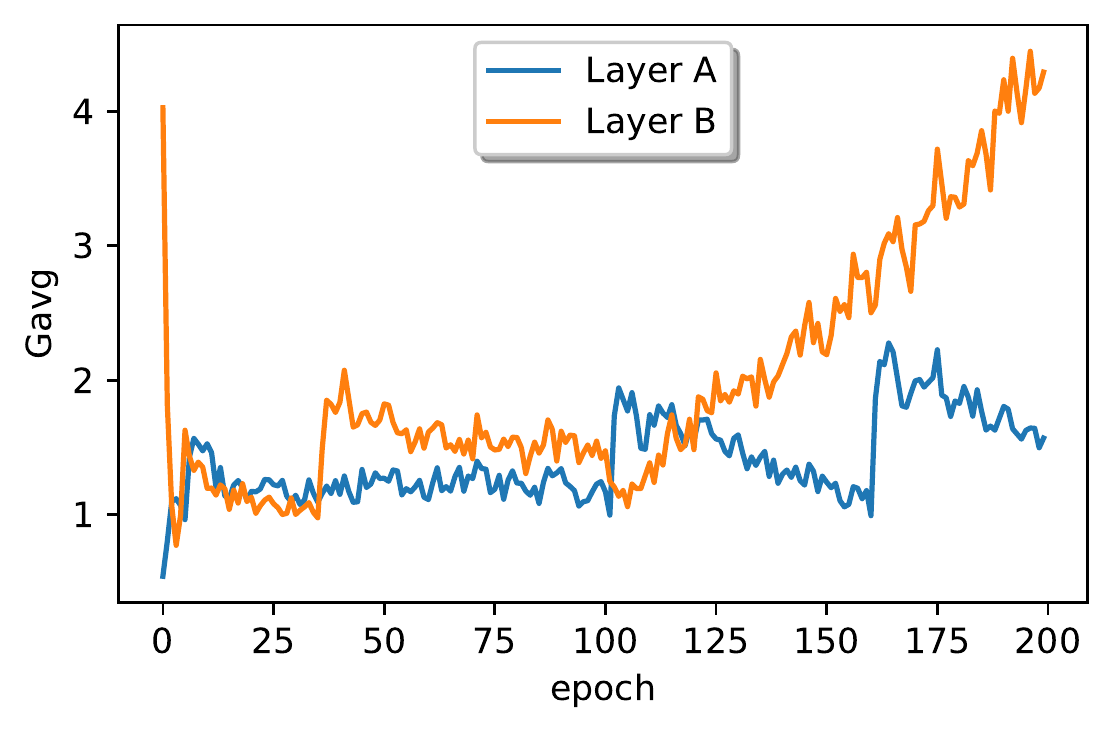}
\caption{Gavg v.s. Epoch for two layers}
\label{fig:gavgvsepoch}
\end{figure}

Fig.\ref{fig:gavgvsepoch} demonstrates the trend of $Gavg$ of two layers in a training with RCT. $T_{min}$ is set to 1.0 in this demo. Layer A starts with a $Gavg$ below $T_{min}$, indicating it suffers quantisation underflow. RCT allocates more bitwidth to lift the $Gavg$ of layer A above the threshold. Layer B is very easy to update at the beginning. Whenever $Gavg$ hits $T_{min}$, RCT allocates more bitwidth to ensure layer B learns effectively. $T_{max}$ is set to $+\infty$ in this demo. It is also possible to use $T_{max}$ to reduce bitwidth for layers that do not suffer much from quantisation underflow.

\begin{figure}[hbt]
\centering
\includegraphics[width=0.6\linewidth]{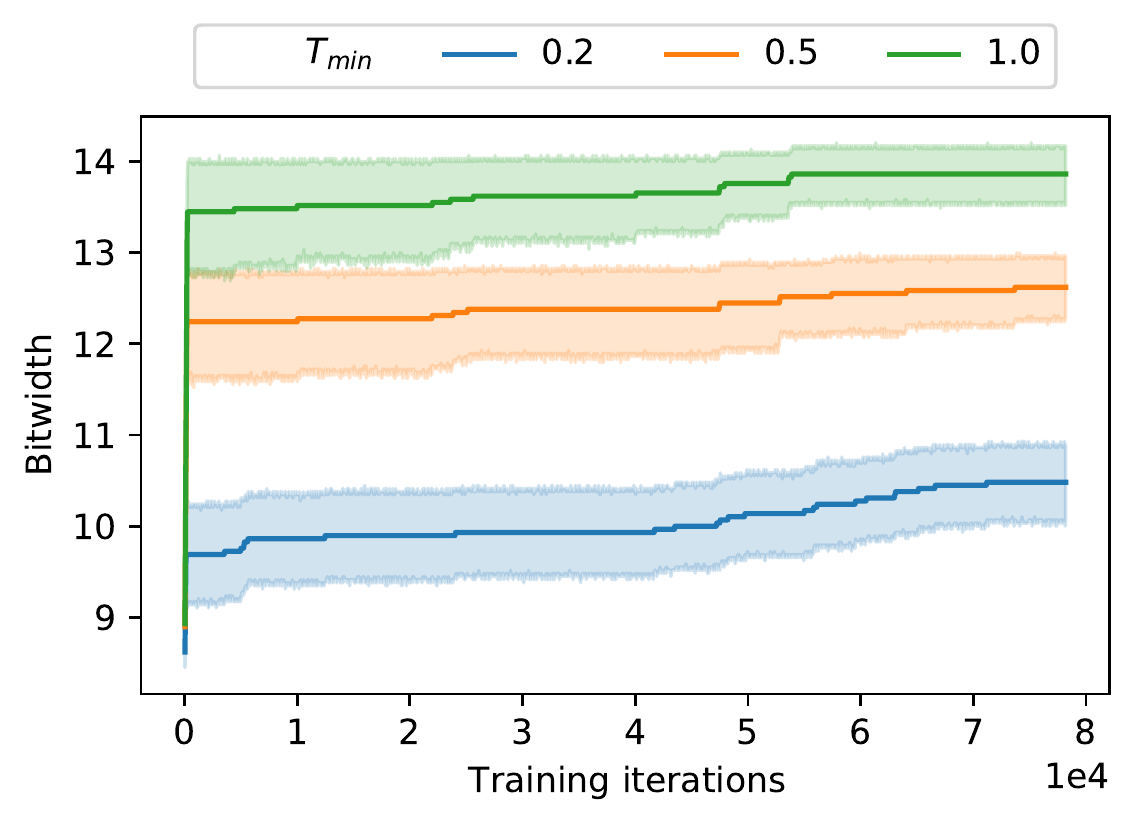}
\caption{Layer-wise Bitwitdh v.s. Epoch for WideResNet-28-10 on CIFAR100}
\label{fig:bitwidthvsepoch}
\end{figure}

Fig.\ref{fig:bitwidthvsepoch} demonstrates the distribution of bitwidth when we train WideResNet-28-10 on CIFAR100 dataset with RCT. The shade represents the 95\% confidence interval. RCT adjusts the bitwidth of the each layer along with the training progress according to $Gavg$. Some layers are trained with lower bitwidth in early epochs and achieve solid energy savings. As the training progress move forward, the average of bitwidth increase gradually, indicates that RCT allocates more bitwidth to the layers with low $Gavg$.

Through Fig.\ref{fig:bitwidthvsepoch} we understand that RCT finds bitwidth configuration on the fly. It serves as a heuristic Neural Architecture Search method for edge devices. We start the training with an initial bitwidth of low precision, which is under-qualified with the target application. RCT will dynamically adjust the bitwidth of the network, and approaches the target accuracy with fewer energy.

\section{Experiments}

\textbf{Quantisation scheme} We apply linear quantisation scheme \cite{jacob2018quantization} to convert floating point numbers to into integers. The scheme is equivalent to an \emph{affine mapping} of integers $q$ to real numbers $r$. The formalised equation is shown as follows $$r=S(q-Z)$$. $S$ and $Z$ represent the scale and zero point of a group of values, or a tensor. All values in a tensor share one $S$ and $Z$. Different tensors have their own $S$ and $Z$. For $k$-bit quantisation, $q$ has $2^k$ possible discrete states. Additionally, we apply Stochastic Rounding \cite{gupta2015deep} to mitigate the errors introduced by quantisation. Unless specified, we use $T_{min}=1.0$ and $T_{max}=100.0$ for bitwidth adjustment policy in all the experiments.


\textbf{Energy Estimation} The energy for GEMM reported in this paper is simulated result. We assume the energy consumption of a quantised GEMM is proportional to that of a 32-bit counterpart. For example, a multiplication between two 16-bit operands consumes 25\% energy compares to a 32-bit multiplication. This approximation is well supported by \cite{horowitz20141}. In this way, the reported energy is less dependent on the spec of a computing platform. We do not compare energy for computation with energy for data movement, because they are closely related to computer architecture as well as manufacturing process. We follow the rule of thumb that off-chip data movement is much more expensive than on-chip computation \cite{horowitz20141}.

\textbf{Computing Infrastructure} We carry out all experiments on a general-purpose workstation, which is equipped with one Intel(R) Core(TM) i9-10900X CPU, 128-Gigabyte DDR4 memory and four NVIDIA Geforce RTX 2080 Ti graphics cards.

\subsection{Image Classification}

We evaluate the RCT with four datasets: CIFAR10, CIFAR100 \cite{krizhevsky2009learning},  SVHN \cite{netzer2011reading} and ImageNet \cite{deng2009imagenet}. Four popular backbones, ResNet-20, ResNet-110 \cite{he2016deep}, MobileNetV2 \cite{sandler2018mobilenetv2} and WideResNet \cite{zagoruyko2016wide}, are included in the experiments. We replace all convolution layers and fully connected layers with our quantised version. Activation is quantised to 8-bit. This choice allows us to produce good energy savings without largely compromising the accuracy. The addition in fully connected layer is not quantised as they only make up for a fraction of the amount of memory and energy for training. We average over three runs of the experiment and report the results.

The quantised layers in these models are all initialised with 8-bit at the beginning of training. The result of training is not sensitive to the initial bitwidth. Please refer to Appendix for related experiments and analysis.

\subsubsection{Savings in Energy}

\begin{figure}[hbt]
\centering
\includegraphics[width=0.6\linewidth]{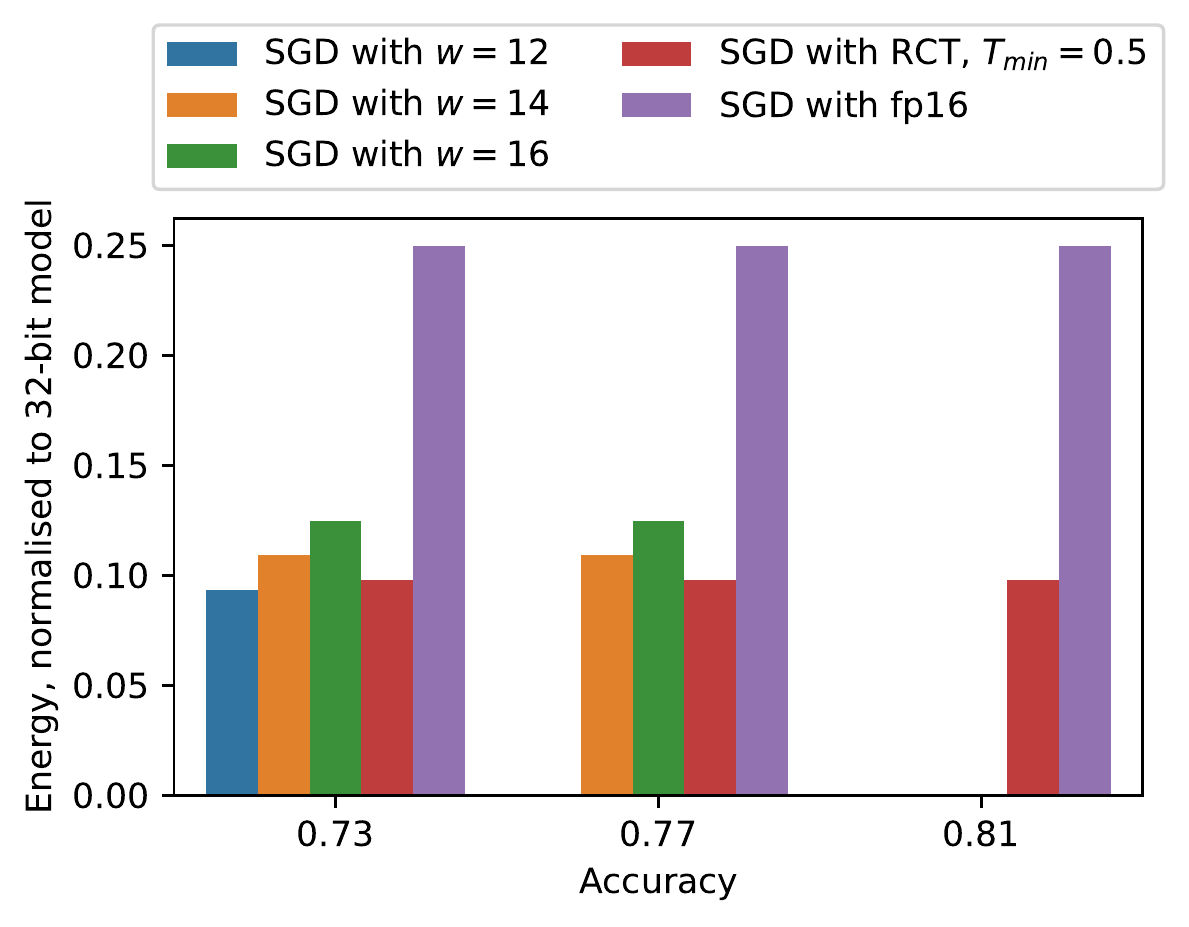}
\caption{Training Energy v.s. Bitwidth for WideResNet-28-10 on CIFAR100}
\label{fig:energyvsbitwidth}
\end{figure}

RCT saves energy when the model can learn with lower precision. The energy here refers to the energy for GEMM in inference. We demonstrate that, with RCT outperforms fixed precision training in accuracy and saves energy compared to floating point training. In this experiment, we use SGD to train WideResNet-28-10 on CIFAR100. We follow the training settings in \cite{devries2017cutout} \footnote{\url{https://github.com/uoguelph-mlrg/Cutout}}. Fixed precision training and floating-point training is included as baselines in this experiment. Fig.\ref{fig:energyvsbitwidth} shows the results of the experiment. 

The SOTA accuracy achieved by SGD with fp32 is 81.59\%. SGD with fp16 achieves similar accuracy with 25\% of energy cost. SGD with RCT is able to achieve the same result with more than 90\% savings in energy, about 15\% more savings than SGD with fp16 does.

SGD with fixed precision $w=14$ and $w=16$ has energy cost that is just a few percent higher than RCT does. However they can only achieve 77\% accuracy, 4\% accuracy loss compared to the 81.59\% SOTA accuracy. Similarly, SGD with $w=12$ consumes about the same energy as RCT does but only reaches 73\% accuracy. The energy saving of fixed precision training comes in the cost of overfitting in the early stage of training and prevent the model from further improvement. We understand that the last few percents improvement in accuracy is usually the most energy consuming part in the whole training progress. This experiments suggests that RCT consumes less energy given the same target accuracy.

In Fig.\ref{fig:energyvsbitwidth}, RCT trains with $T_{min}=0.5$. This hyper-parameter is application specific, which can be used as a trade-off between accuracy, memory and energy. Next we investigate the trade-off in the experiment. We train WideResNet-28-10 on CIFAR100 and ResNet-18 on CIFAR10. We follow the training setting in \cite{devries2017cutout}.

\begin{figure}[htb]
     \centering
     \begin{subfigure}[b]{0.49\linewidth}
         \centering
         \includegraphics[width=\linewidth]{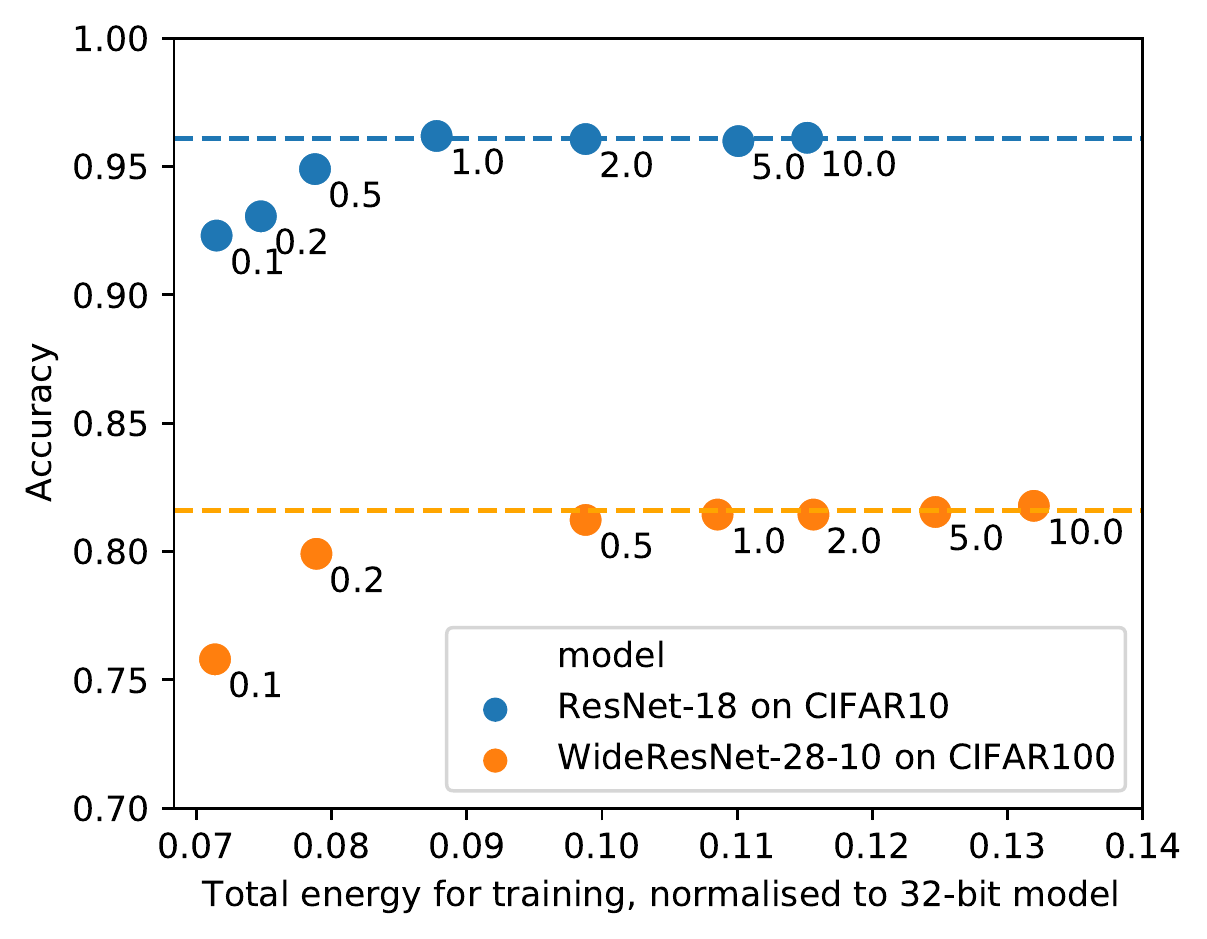}
         \caption{Energy consumption for GEMM v.s. test accuracy}
         \label{fig:energy_accuracy}
     \end{subfigure}
     \hfill
     \begin{subfigure}[b]{0.49\linewidth}
         \centering
         \includegraphics[width=\linewidth]{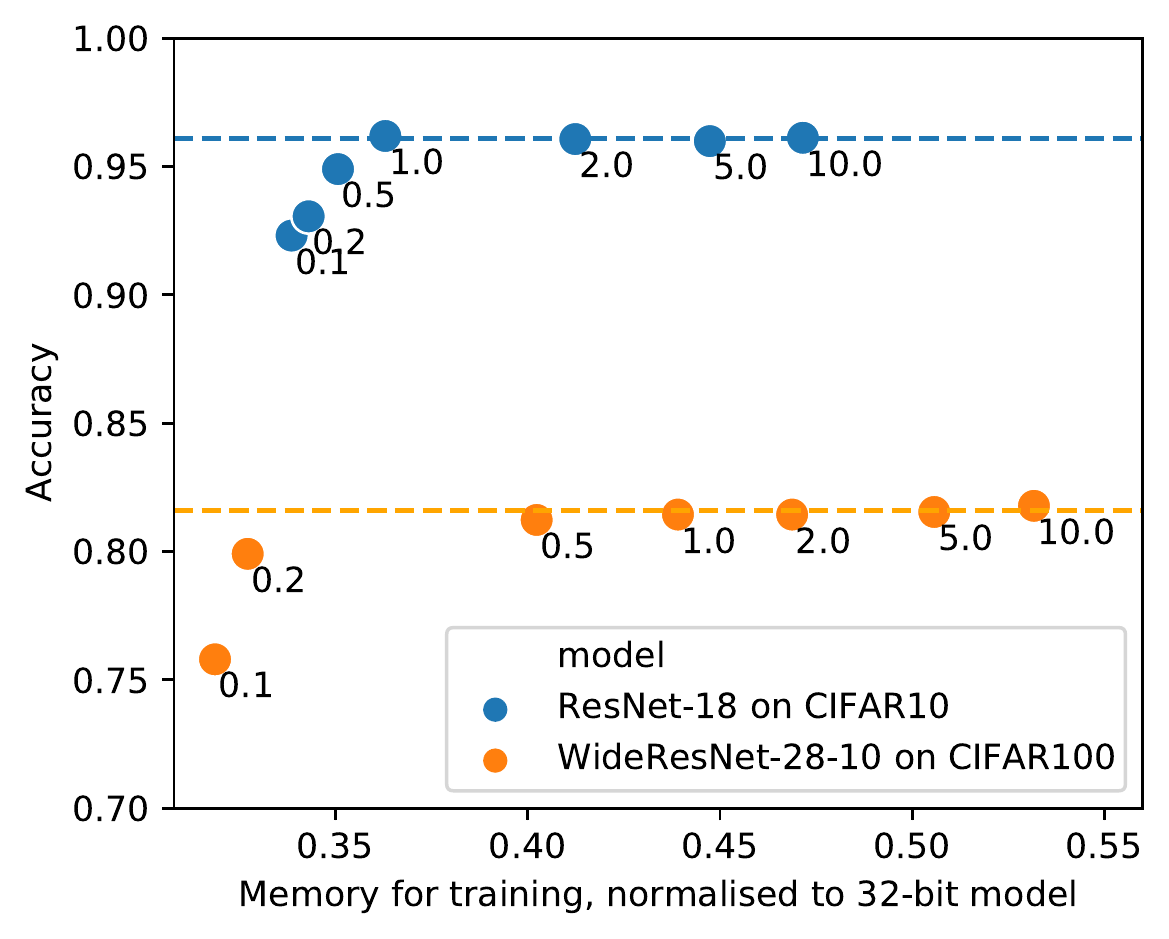}
         \caption{Memory requirements for parameters v.s. test accuracy}
         \label{fig:memory_accuracy}
     \end{subfigure}

        \caption{Image applications. $T_{min}$ plays trade-off between scores and resource requirements. (a) Accuracy v.s. Memory for model parameters, normalised. (b) Accuracy v.s. Energy for GEMM, normalised}
        \label{fig:img_resource_vs_score}
\end{figure}


In Fig.\ref{fig:energy_accuracy}, each point represents the results of a 150-epoch training using RCT. Threshold $T_{min}$ for metric $Gavg$ is ranging from 0.1 to 10. The dashed lines represents the SOTA accuracy. From the figure we know that for both models runs with $T_{min}\geq1.0$ result in accuracy very close to the SOTA accuracy. $T_{min}=0.5$ for ResNet-18 on CIFAR10 produces one more percent energy savings in the cost of accuracy loss of less than 1\%. Overall, $T_{min}=0.5$ for both models provides at least 90\% energy savings with accuracy loss no more than 1\%.

\subsubsection{Savings in Memory}

RCT only keeps quantised model parameters. The memory requirements for model parameters in training is reduced. Next we investigate the memory savings for model parameters achieved by RCT. We do not include memory for activation into the experiment, because this is not the focus of our method. One can use micro-batch technique \cite{huang2019gpipe} such that the memory requirements for activation is trivial comparing with model parameters.


The setting for Fig.\ref{fig:memory_accuracy} is similar to Fig.\ref{fig:energy_accuracy} except that x axis represents memory requirements for training, normalised to 32-bit models. From the figure we know that for both models runs with $T_{min}\geq1.0$ result in accuracy very close to the SOTA accuracy. $T_{min}=0.5$ for ResNet-18 on CIFAR10 produces one more percent memory savings in the cost of accuracy loss below 1\%. Overall, $T_{min}=0.5$ for both models provides at least 65\% memory savings with accuracy loss no more than 1\%.

\subsection{NLP tasks}

We evaluate the RCT with Natural Language Processing (NLP) tasks. Pre-trained Transformer \cite{vaswani2017attention} based language models, such as BERT \cite{devlin2018bert}, have shown great improvement in many NLP tasks. In this experiment, we include text classification task set GLUE \cite{wang2018glue} and the question-answering task, SQuADv1.1 \cite{rajpurkar2016squad}. We employ BERT\footnote{bert-base-cased for GLUE, bert-base-uncased for SQuADv1.1} as the backbone. We replace all embedding layers and fully connected layers with our quantised version. The precision of weights in these layers are all initialised as 8-bit at the beginning of training. Unless specified, $(T_{min}, T_{max})$ is set to $(1,0, 100.0)$ for all experiments. Activation is quantised to 8-bit. The addition in fully connected layer is not quantised as they only make up for a fraction of the amount of memory and energy for training. We set batch size to 16 for GLUE and 12 for SQuADv1.1 and follow the rest of data processing and training methods presented in open-source library transformers \footnote{\url{https://github.com/huggingface/transformers/tree/v4.1.1}}. We adopt random seeds in the experiment such that the reproducibility of the results is guaranteed.

\begin{table}[htb]
\caption{NLP performance}
\centering
\begin{tabular}{c c c c c}
\hline
Task      & Metric         & fp32  & RCT   & Bitwidth       \\ \hline
CoLA      & Matthew's corr & 56.53 & \textbf{57.29} & 18.0    \\ 
MRPC      & F1             & 88.85 & \textbf{90.20} & 17.9    \\ 
QNLI      & Accuracy       & 90.66 & \textbf{90.87} & 17.7    \\ 
QQP       & F1             & 87.49 & 86.74 & 17.6    \\ 
RTE       & Accuracy       & 65.70 & 65.34 & 18.8    \\ 
SST-2     & Accuracy       & 92.32 & 91.97 & 18.2    \\ 
STS-B     & Person corr.   & 88.64 & \textbf{88.78} & 17.9    \\ \hline
SQuADv1.1     & F1             & 88.46 & 87.62 & 18.2    \\ \hline
\end{tabular}
\label{tab:comparison_nlp}
\end{table}

We report our NLP experiement results in Table \ref{tab:comparison_nlp}. The fp32 column is the score of a floating-point SGD method, which serves as a baseline. The RCT column shows the scores achieved by RCT method. RCT produces scores that are very close to the SOTA scores by the fp32 baseline method. The last column shows the averaged bitwidth of the quantised model at the end of RCT training.

\begin{figure}[hbt]
\centering
\includegraphics[width=\linewidth]{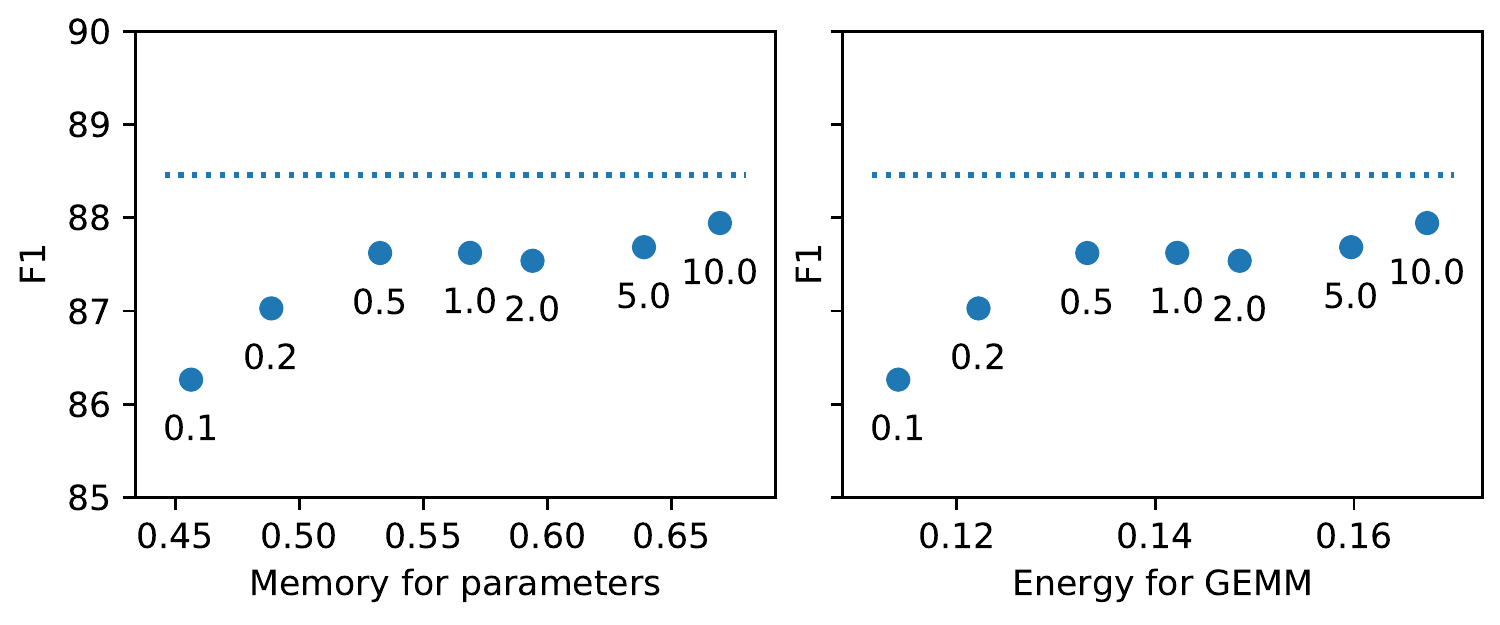}
\caption{BERT on SQuADv1.1: $T_{min}$ plays trade-off between scores and resource requirements. Left: F1 v.s. Memory for model parameters, normalised. Right: F1 v.s. Energy for GEMM, normalised}
\label{fig:bert_squad_resource_vs_score}
\end{figure}

Hyper-parameter $T_{min}$ plays important roles in trade-off between the score and the resource required by the training process. Figure \ref{fig:bert_squad_resource_vs_score} shows the relation between them for BERT on SQuADv1.1.

In Figure \ref{fig:bert_squad_resource_vs_score}, each point represents the result of a 2-epoch training process. The horizontal dotted line represents the score achieved by fp32 baseline. When $T_{min}\geq0.5$ the loss in score caused by RCT is always within 1\%. When  $T_{min}<0.5$ the score starts to drop gradually. This Figure together with Figure \ref{fig:memory_accuracy} and Figure \ref{fig:energy_accuracy} suggests that $T_{min}=0.5$ is a reasonable choice for various applications.

\subsection{Comparison with Others}

\begin{table*}[htb]
\caption{Memory and Energy Comarison}
\begin{center}
\begin{threeparttable}
\begin{tabular}{ c c c c c c }
\hline
Method                              & Model            & Dataset  & Accuracy loss \tnote{a} & Memory \tnote{b} & Energy \tnote{c} \\ \hline
XNOR-Net \cite{rastegari2016xnor}   & AlexNet          & ImageNet & 12.4\%            & 32+1 bits       & - \\
TWN \cite{li2016ternary}            & ResNet-18        & ImageNet & 3.6\%             & 32+3 bits       & 2.73 \\
QNN \cite{banner2018scalable}       & ResNet-18        & ImageNet & $\leq1\%$         & 32+8 bits       & 3.125 \\
WAGEUBN \cite{yang2020training}     & ResNet-18        & ImageNet & 3.9\%             & 24+8 bits       & 2.5 \\
MP \cite{micikevicius2017mixed}     & ResNet-50        & ImageNet & $\leq1\%$         & 32+16 bits      & - \\
PG \cite{zhang2020precision}        & ShuffleNetV2     & ImageNet & $\leq1\%$         & 32 bits         & 2.48 \\
Q8BERT \cite{zafrir2019q8bert}      & BERT             & GLUE     & $\leq1\%$         & 32+8 bits       & $\sim$2.22 \\
Q8BERT \cite{zafrir2019q8bert}      & BERT             & SQuADv1.1    & $\leq1\%$         & 32+8 bits       & 2.20 \\
Q-BERT \cite{shen2020q}             & BERT             & SQuADv1.1    & $2.3\%$           & 32+3.7 bits     & 1.96 \\
BinaryBERT \cite{bai2020binarybert} & BERT             & SQuADv1.1    & $\leq1\%$         & 32+1 bits       & 1.81 \\
\hline
\multirow{7}{*}{RCT}                & MobileNetV2      & ImageNet & \multirow{7}{*}{$\leq1\%$} 
                                                                                      & 12.1 bits       & -  \\
                                    & ShuffleNetV2     & ImageNet &                   & 12.9 bits       & 1  \\
                                    & ResNet-18        & ImageNet &                   & 12.8 bits       & 1  \\
                                    & ResNet-18        & CIFAR10  &                   & 11.2 bits       & -  \\
                                    & WideResNet-28-10 & CIFAR100 &                   & 10.6 bits       & -  \\
                                    & BERT             & GLUE     &                   & $\sim$18 bits   & 1  \\
                                    & BERT             & SQuADv1.1    &                   & 18.2 bits       & 1  \\

\hline
\end{tabular}
\begin{tablenotes}
    \item[a] The metrics for different tasks: Top1 accuracy for ImageNet, accuracy for CIFAR-10, F1 for SQuADv1.1, please refer to Table \ref{tab:comparison_nlp} for the metrics for GLUE task set.
    \item[b] Memory for model parameters in training. We use averaged bitwidth as a model-independent indicator for memory requirements. For RCT, we report weighted average at the end of training.
    \item[c] Energy for moving model parameters between off-chip memory and processor, normalised to the counterpart of RCT.

\end{tablenotes}
\end{threeparttable}
\end{center}
\label{tab:comparison}
\end{table*}

Table \ref{tab:comparison} shows the comparison between RCT and other representative training methods. Many of these methods involve two copies of model parameters  during training, e.g., an fp32 copy and a quantised copy, thus have the form of $32+x$ bits in the memory column. RCT only involves one copy of model parameters, which has a better chance to fit into on-chip memory of a processor and stay inside throughout the training process.

In the case where the model parameters do not fit into on-chip memory, RCT still saves energy for data movement. We estimate that the energy for data movement is linearly proportional to the size of memory for model parameters. For example the energy cost by the training method for QNN is 3.125 times of what RCT requires, because $\frac{32+8}{12.8}=3.125$. According to \cite{horowitz20141}, DRAM operations, i.e., off-chip memory read/write, could be $10^2-10^4$ times more expensive than on-chip operations do.

\section{Discussion}

\subsection{Resource Efficient Hardware}

Resource efficient hardware techniques are necessary for fully exploiting the benefit of RCT. The techniques here refer to the ability that a processing unit saves energy and memory according to the bit-width of operands, meaning that the fewer bit-width, the less consumption in energy and memory. There has been a growing attention towards these techniques. For example, the computation time and energy of BISMO \cite{judd2016stripes, umuroglu2018bismo} is linearly proportional to the bitwidth of input operands. Bit Fusion \cite{sharma2018bit} is a bit-flexible accelerator, that constitutes an array of bit-level processing elements that dynamically fuse to match the bitwidth of individual DNN layers. For VGG-7 on CIFAR10, Bit Fusion consumes 895 milliwatts, which is two orders of magnitude lower than a 250-Watt Titan Xp that uses 8-bit computations.

There are also studies for efficient memory in variable bit-width. The representative research ShapeShifter \cite{lascorz2019shapeshifter} suggested that a smart encoding and decoding scheme can mitigate memory alignment issues and can saves up to 80\% bandwidth/memory. More importantly, it can seamlessly cooperate with BISMO and Bit Fusion architecture. The overhead introduced by this technique is trivial compares to the benefits and opportunity it brings to us. Hardware with such techniques facilitates in-situ learning for resource constrained edge AI/ML.

\subsection{Distribution of Bitwidth}

\begin{figure}[hbt]
\centering
\includegraphics[width=0.6\linewidth]{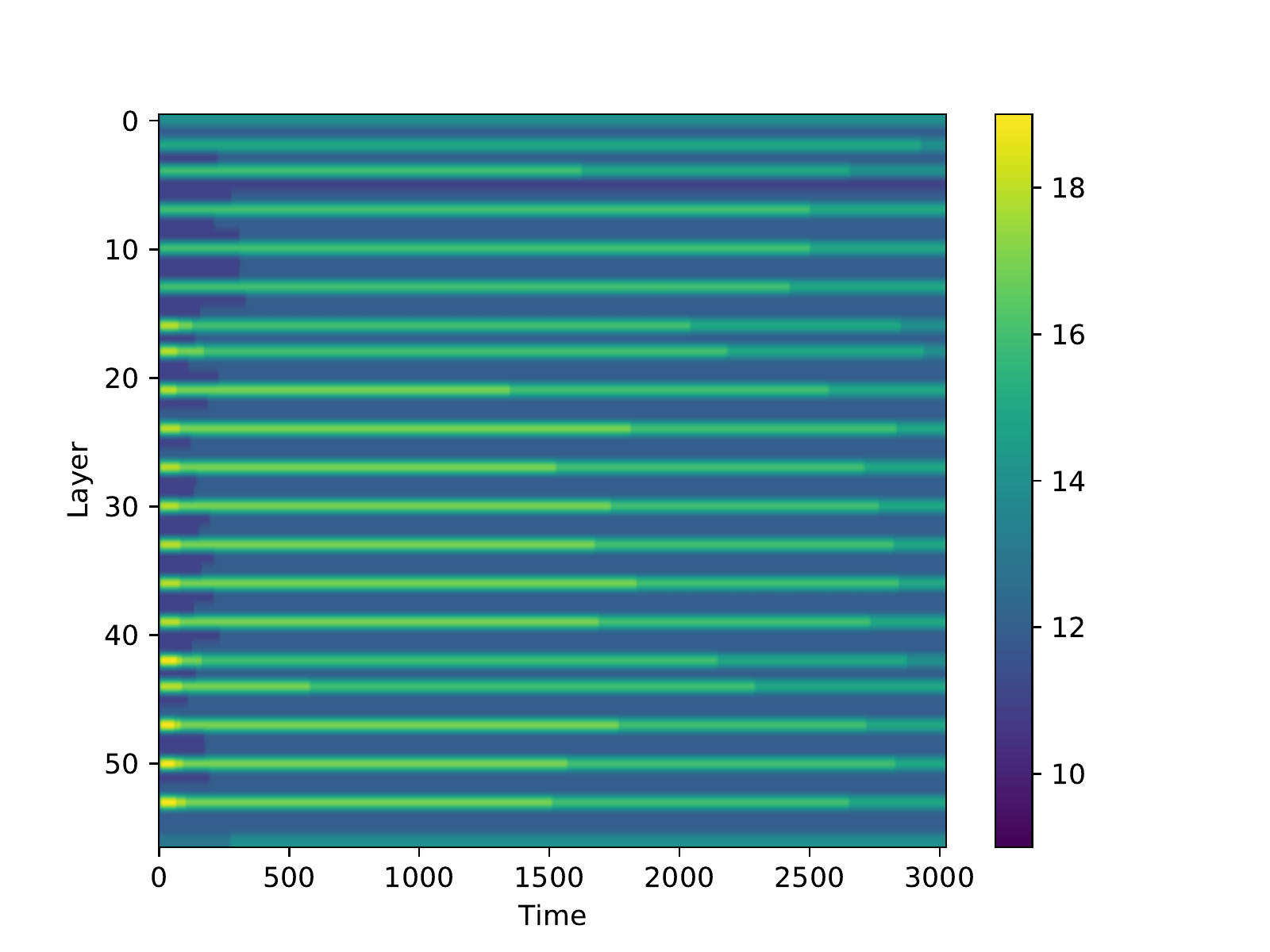}
\caption{ShuffleNetV2 on ImageNet, history of bitwidth. x-axis stands for time, y-axis stands for different layers. First layer is at the top. Each time tick represents 100 steps}
\label{fig:snv2_imagenet_bw_history_demo}
\end{figure}

Figure \ref{fig:snv2_imagenet_bw_history_demo} shows the distribution of bitwidth for ShuffleNetV2 on ImageNet. The light color implies quantisation underflow happens more often in the branch\_main.3 of each basic unit at the beginning of training. In the second half of the training process, the bitwidth of these layers drop slightly, which suggests underflow happens less often. Other models and datasets have rather different distribution of bitwidth. Please refer to Appendix for more details.

\subsection{Bitwidth v.s. Batch Size}

Larger batch size leads to larger gradients and less frequent quantisation underflow. That means a model can learn effectively with lower bitwidth. Please refer to Appendix for detailed experiment on this feature of RCT. 

In many edge applications, the batch size of training data is usually quite limited. \emph{micro-batching} \cite{huang2018gpipe} is one of methods that can mitigate the limitation in batch size. It is also known as \emph{Gradient Accumulation}, which is widely adopted by NLP research community. This method accumulates gradients from multiple batches before carrying out model update. With this technique, RCT can have larger gradients and lower quantisation underflow, which leads to models with lower bitwidth.

\section{Conclusion}

In-situ training for resource constrained edge AI is challenging. We propose Resource Constrained Training method that saves both energy cost and memory usage for training. RCT evaluates how effective each layer learns with its current precision. Based on the evaluation, RCT performs per-layer bitwidth adjustments dynamically to make sure the model learns effective throughout the training process. We evaluate RCT with image classification and NLP tasks. Results suggest that RCT effectively saves memory and energy for training, without compromising the accuracy of a model.

\bibliographystyle{unsrt}  
\bibliography{reference}

\clearpage

\appendix
\appendixpage
\addappheadtotoc

\section{Initialisation of Bitwidth}

\begin{figure}[hbt!]
\centering
\includegraphics[width=0.6\linewidth]{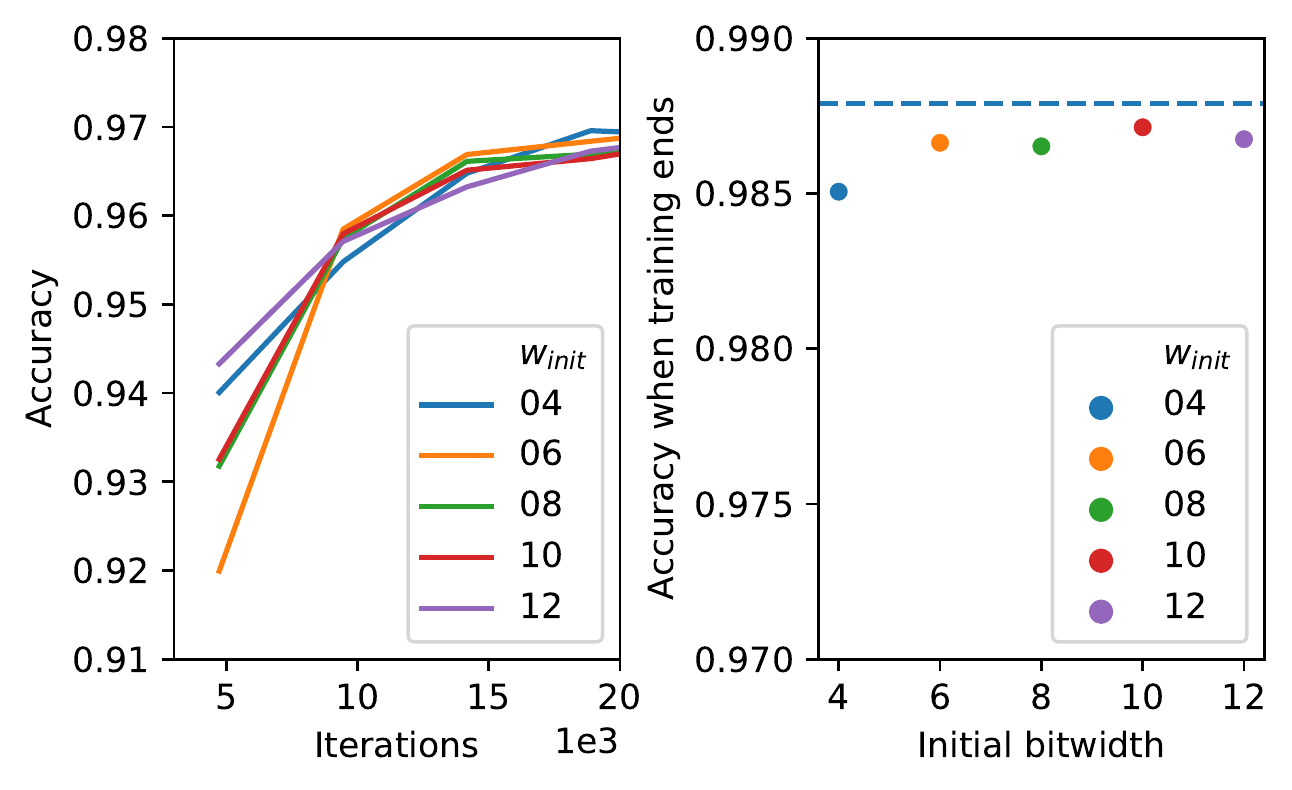}
\caption{Accuracy v.s. different initial bitwidth}
\label{fig:bwinit}
\end{figure}

The accuracy at the end of training is not sensitive to the setting of the initial bitwidth. To prove this, we use RCT to train WideResNet-28-10 on SVHN with different settings of the initial bitwidth, ranging from 4-bit to 12-bit. For the rest of data augmentation and training setting we follow \cite{zagoruyko2016wide}. The results is shown in Fig.\ref{fig:bwinit}.

The left part of Fig.\ref{fig:bwinit} demonstrates the accuracy curve at the beginning of the training. Models with different initial bitwidth has different starting accuracy at the first first epoch. The divergence gradually narrows, because RCT is allocating more bitwidth to models with fewer initial bitwidth. The right part of the figure shows the accuracy at the end of the training achieved by these models. All the points are very close to the dashed line, which represents the accuracy achieved by floating-point training. This experiment suggests that the RCT is not sensitive to the initial bitwidth. Therefore we will use initial bitwdith of 8-bit for the rest of the experiments

\begin{figure}[hbt!]
\centering
\includegraphics[width=0.6\linewidth]{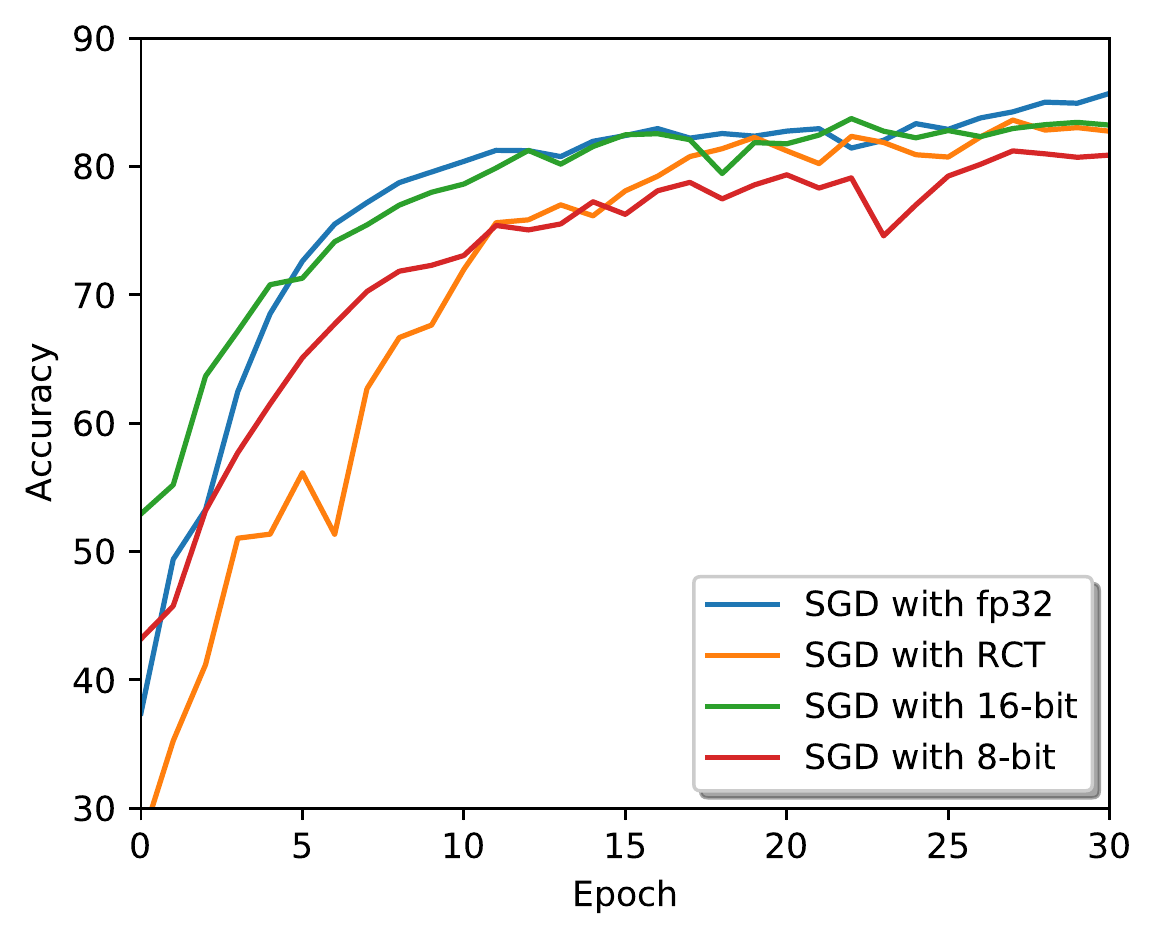}
\caption{Test Accuracy v.s. Epoch for ResNet20 on CIFAR10}
\label{fig:accvsepoch}
\end{figure}

We then demonstrate that, starting with low-precision model, RCT can catch up with the progress of floating-point training. We follow the training settings in \cite{he2016deep} Fig.\ref{fig:accvsepoch} shows the first few epochs of the training progress of ResNet-20 on CIFAR10.

FP32 and 16-bit model has the steepest learning curves among all, as they do not suffer much from quantisation underflow problem. The curve of 8-bit model does not climb as fast as FP32 and 16-bit does. An investigation into the training statistics shows that $Gavg$ of the all layers drop from the scale of 1 down to $1e-1$ within the first 50 epoch, which indicates quantisation underflow happens model wide and significantly slows down the training process of the 8-bit model. On comparison, RCT starts with a model initialised with 8-bit weights. Its training curve starts with lower accuracy at the beginning. It overtakes the 8-bit and catch up with 16-bit and FP32.

\section{Bitwidth v.s. Batch Size}

\begin{figure}[hbt]
\centering
\includegraphics[width=0.6\linewidth]{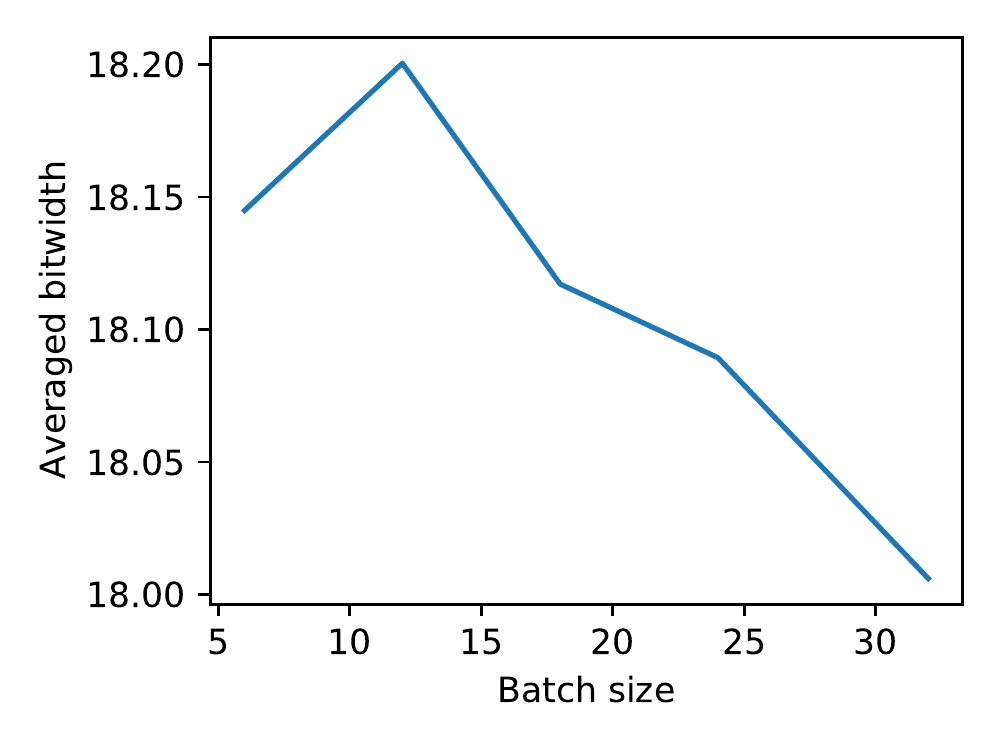}
\caption{BERT on SQuADv1.1, batch size v.s. bitwidth. x-axis stands for batch size, y-axis is the averaged bitwidth at the end of RCT training.}
\label{fig:ber_squad_bs_vs_bw}
\end{figure}

Figure \ref{fig:ber_squad_bs_vs_bw} shows the relation between batch size and bitwdith for BERT on SQuADv1.1. This figure suggests that larger batch size leads to lower bitwidth. This is because quantisation underflow happens less often if the gradient is larger. The batch size for language model is much smaller than that for CNN models. For example, batch size is 12 for BERT on SQuADv1.1 and 1024 for ShuffleNetV2 on ImageNet. This explain why the averaged bitwidth for language model is a few bits higher than that for CNN models.

\section{Distribution of Bitwidth Over Time}

\subsection{ResNet-18 on Imagenet}

\begin{figure}[hbt]
\centering
\includegraphics[width=0.6\linewidth]{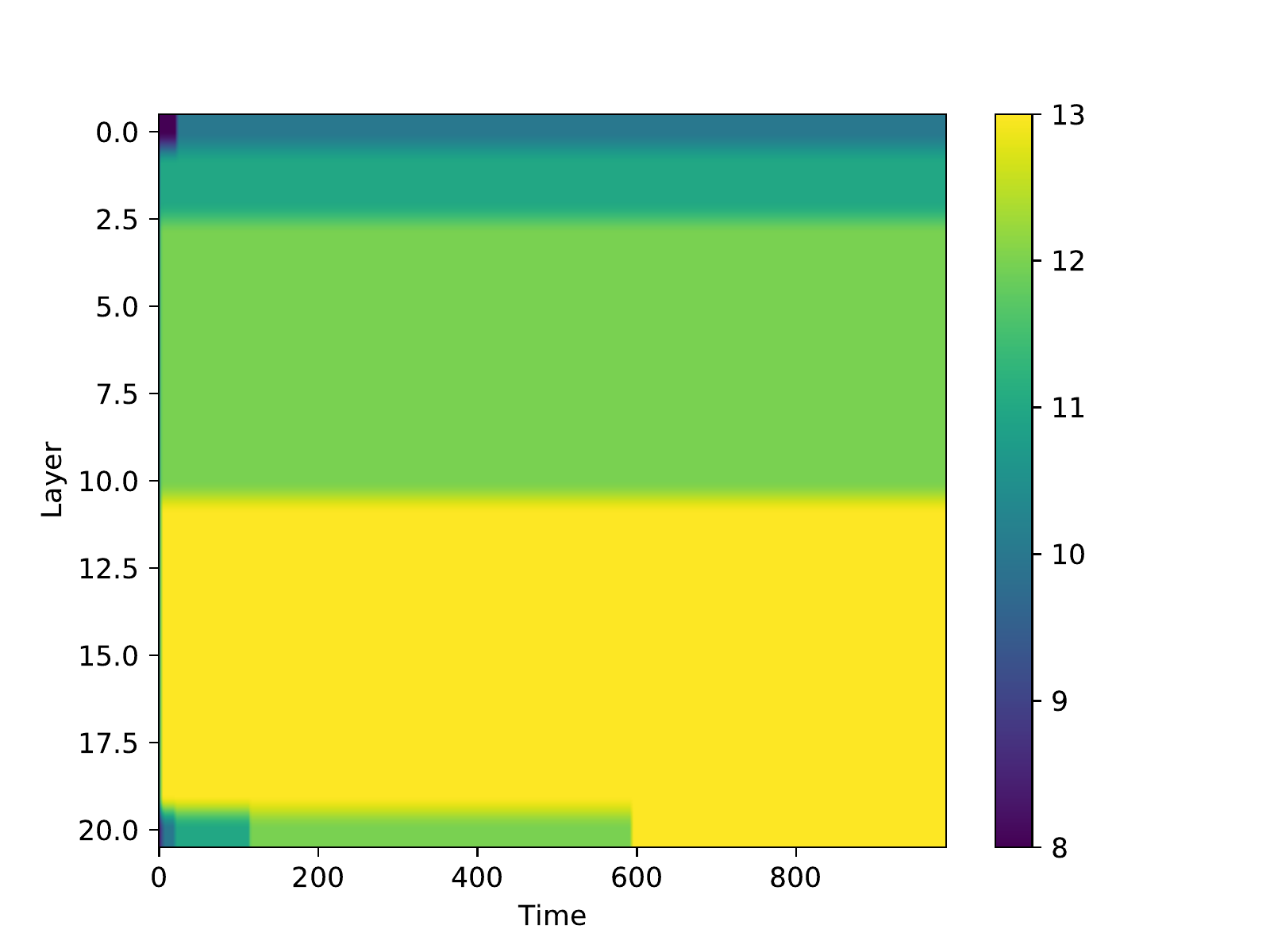}
\caption{ResNet-18 on ImageNet, history of bitwidth. x-axis stands for time, y-axis stands for different layers. First layer is at the top. Each time tick represents 500 steps}
\label{fig:rn18_imagenet_bw_history}
\end{figure}

\begin{center}
\tablefirsthead{
\multicolumn{1}{l}{Layer name} & \multicolumn{1}{l}{Bitwidth} \\ \hline
}
\tablehead{
\multicolumn{2}{l}{\small\sl continued from previous page}\\ \hline
\multicolumn{1}{l}{Layer name} & \multicolumn{1}{l}{Bitwidth} \\ \hline
}
\tabletail{%
\hline
\multicolumn{2}{r}{\small\sl continued on next page}\\
}
\tablelasttail{\hline}
\tablecaption{ResNet-18 on Imagenet, bitwidth at last step}
\begin{supertabular}{l l}
conv1	&	10.0	\\
layer1.0.conv1	&	11.0	\\
layer1.0.conv2	&	11.0	\\
layer1.1.conv1	&	12.0	\\
layer1.1.conv2	&	12.0	\\
layer2.0.conv1	&	12.0	\\
layer2.0.conv2	&	12.0	\\
layer2.0.downsample.0	&	12.0	\\
layer2.1.conv1	&	12.0	\\
layer2.1.conv2	&	12.0	\\
layer3.0.conv1	&	12.0	\\
layer3.0.conv2	&	13.0	\\
layer3.0.downsample.0	&	13.0	\\
layer3.1.conv1	&	13.0	\\
layer3.1.conv2	&	13.0	\\
layer4.0.conv1	&	13.0	\\
layer4.0.conv2	&	13.0	\\
layer4.0.downsample.0	&	13.0	\\
layer4.1.conv1	&	13.0	\\
layer4.1.conv2	&	13.0	\\
fc	&	13.0	\\
\label{tab:rn18_imagenet_bw_last_step}\\
\end{supertabular}
\end{center}

\subsection{ShuffleNetV2 on Imagenet}

\begin{figure}[hbt]
\centering
\includegraphics[width=0.6\linewidth]{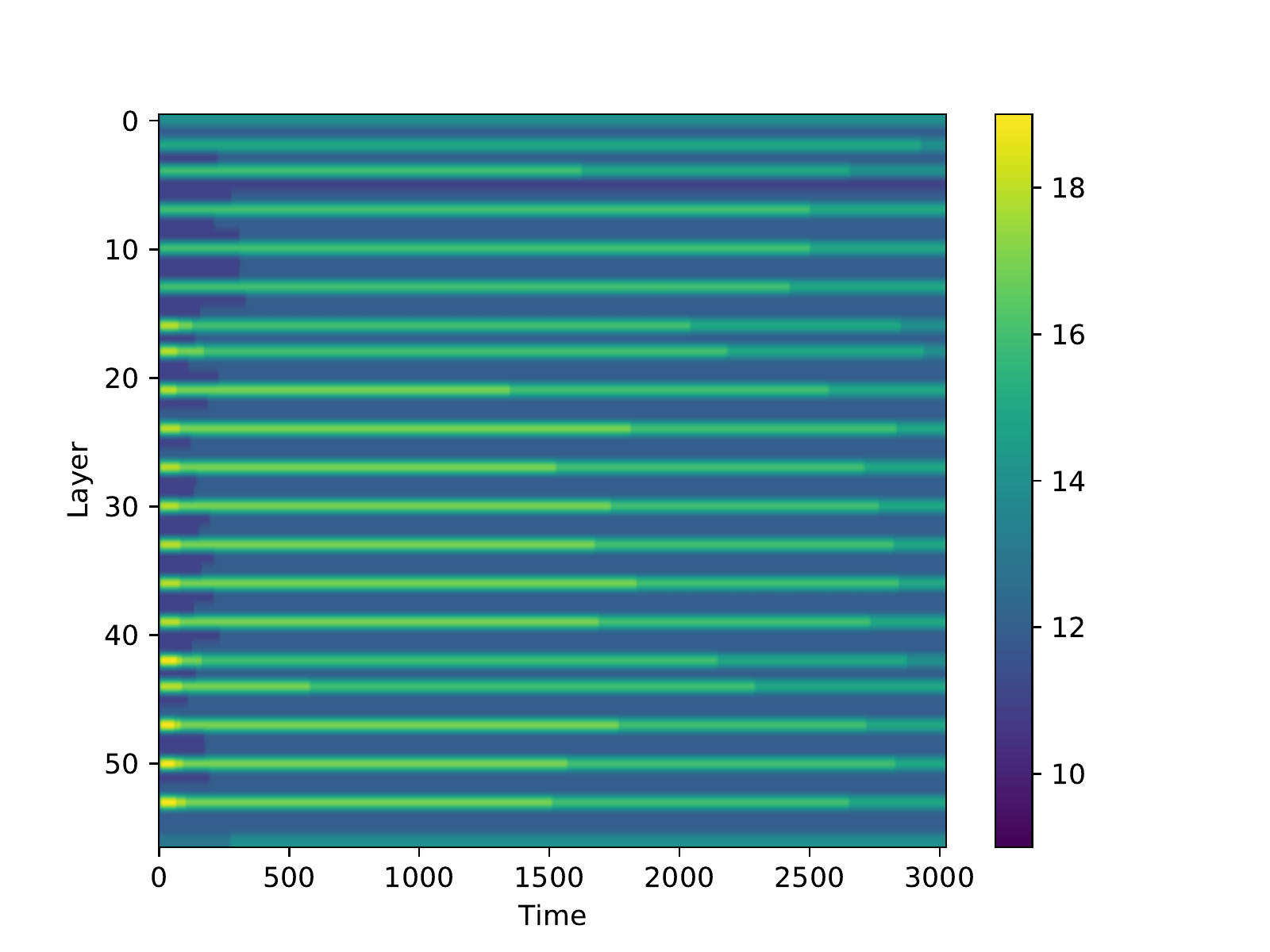}
\caption{ShuffleNetV2 on ImageNet, history of bitwidth. x-axis stands for time, y-axis stands for different layers. First layer is at the top. Each time tick represents 100 steps}
\label{fig:snv2_imagenet_bw_history}
\end{figure}

\begin{center}
\tablefirsthead{
\multicolumn{1}{l}{Layer name} & \multicolumn{1}{l}{Bitwidth} \\ \hline
}
\tablehead{
\multicolumn{2}{l}{\small\sl continued from previous page}\\ \hline
\multicolumn{1}{l}{Layer name} & \multicolumn{1}{l}{Bitwidth} \\ \hline
}
\tabletail{%
\hline
\multicolumn{2}{r}{\small\sl continued on next page}\\
}
\tablelasttail{\hline}
\tablecaption{ShuffleNetV2 on ImageNet, bitwidth at last step}
\begin{supertabular}{l l}
first\_conv.0	&	14.0	\\
features.0.branch\_main.0	&	12.0	\\
features.0.branch\_main.3	&	14.0	\\
features.0.branch\_main.5	&	12.0	\\
features.0.branch\_proj.0	&	14.0	\\
features.0.branch\_proj.2	&	11.0	\\
features.1.branch\_main.0	&	12.0	\\
features.1.branch\_main.3	&	15.0	\\
features.1.branch\_main.5	&	12.0	\\
features.2.branch\_main.0	&	12.0	\\
features.2.branch\_main.3	&	15.0	\\
features.2.branch\_main.5	&	12.0	\\
features.3.branch\_main.0	&	12.0	\\
features.3.branch\_main.3	&	15.0	\\
features.3.branch\_main.5	&	12.0	\\
features.4.branch\_main.0	&	12.0	\\
features.4.branch\_main.3	&	14.0	\\
features.4.branch\_main.5	&	12.0	\\
features.4.branch\_proj.0	&	14.0	\\
features.4.branch\_proj.2	&	12.0	\\
features.5.branch\_main.0	&	12.0	\\
features.5.branch\_main.3	&	15.0	\\
features.5.branch\_main.5	&	12.0	\\
features.6.branch\_main.0	&	12.0	\\
features.6.branch\_main.3	&	15.0	\\
features.6.branch\_main.5	&	12.0	\\
features.7.branch\_main.0	&	12.0	\\
features.7.branch\_main.3	&	15.0	\\
features.7.branch\_main.5	&	12.0	\\
features.8.branch\_main.0	&	12.0	\\
features.8.branch\_main.3	&	15.0	\\
features.8.branch\_main.5	&	12.0	\\
features.9.branch\_main.0	&	12.0	\\
features.9.branch\_main.3	&	15.0	\\
features.9.branch\_main.5	&	12.0	\\
features.10.branch\_main.0	&	12.0	\\
features.10.branch\_main.3	&	15.0	\\
features.10.branch\_main.5	&	12.0	\\
features.11.branch\_main.0	&	12.0	\\
features.11.branch\_main.3	&	15.0	\\
features.11.branch\_main.5	&	12.0	\\
features.12.branch\_main.0	&	12.0	\\
features.12.branch\_main.3	&	14.0	\\
features.12.branch\_main.5	&	12.0	\\
features.12.branch\_proj.0	&	15.0	\\
features.12.branch\_proj.2	&	12.0	\\
features.13.branch\_main.0	&	12.0	\\
features.13.branch\_main.3	&	15.0	\\
features.13.branch\_main.5	&	12.0	\\
features.14.branch\_main.0	&	12.0	\\
features.14.branch\_main.3	&	15.0	\\
features.14.branch\_main.5	&	12.0	\\
features.15.branch\_main.0	&	12.0	\\
features.15.branch\_main.3	&	15.0	\\
features.15.branch\_main.5	&	12.0	\\
conv\_last.0	&	12.0	\\
classifier.0	&	14.0	\\
\label{tab:snv2_imagenet_bw_last_step}\\
\end{supertabular}
\end{center}

\subsection{BERT on MRPC}

\begin{figure}[hbt]
\centering
\includegraphics[width=0.6\linewidth]{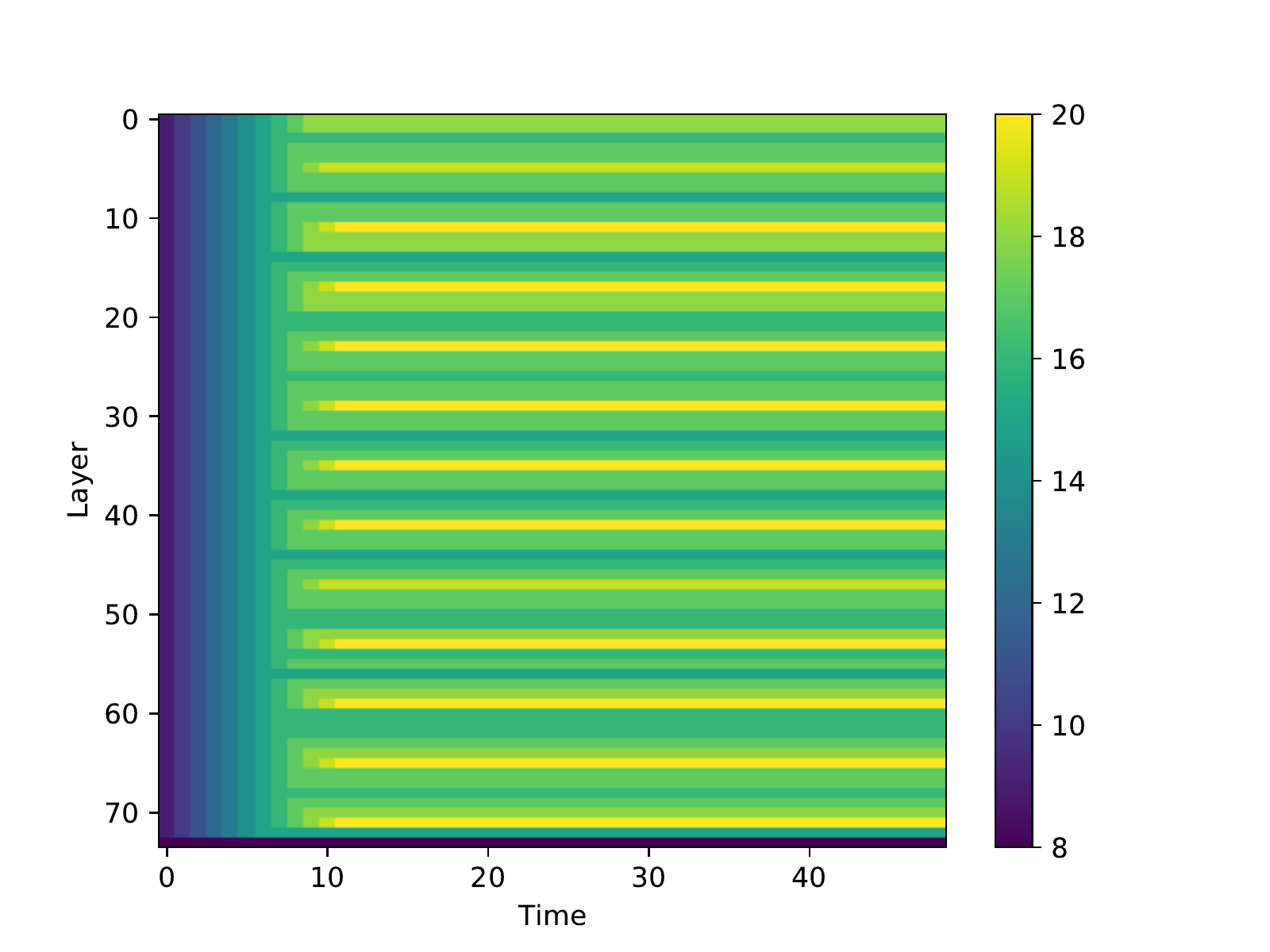}
\caption{BERT on MRPC, history of bitwidth. x-axis stands for time, y-axis stands for different layers. First layer is at the top. Each time tick represents 23 steps}
\label{fig:bert_mrpc_bw_history}
\end{figure}

\begin{center}
\tablefirsthead{
\multicolumn{1}{l}{Layer name} & \multicolumn{1}{l}{Bitwidth} \\ \hline
}
\tablehead{
\multicolumn{2}{l}{\small\sl continued from previous page}\\ \hline
\multicolumn{1}{l}{Layer name} & \multicolumn{1}{l}{Bitwidth} \\ \hline
}
\tabletail{%
\hline
\multicolumn{2}{r}{\small\sl continued on next page}\\
}
\tablelasttail{\hline}
\tablecaption{BERT on MRPC, bitwidth at last step}
\begin{supertabular}{l l}
bert.encoder.layer.0.attention.self.query	&	18.0    \\
bert.encoder.layer.0.attention.self.key	&	18.0    \\
bert.encoder.layer.0.attention.self.value	&	16.0    \\
bert.encoder.layer.0.attention.output.dense	&	17.0    \\
bert.encoder.layer.0.intermediate.dense	&	17.0    \\
bert.encoder.layer.0.output.dense	&	19.0    \\
bert.encoder.layer.1.attention.self.query	&	17.0    \\
bert.encoder.layer.1.attention.self.key	&	17.0    \\
bert.encoder.layer.1.attention.self.value	&	15.0    \\
bert.encoder.layer.1.attention.output.dense	&	17.0    \\
bert.encoder.layer.1.intermediate.dense	&	17.0    \\
bert.encoder.layer.1.output.dense	&	20.0    \\
bert.encoder.layer.2.attention.self.query	&	18.0    \\
bert.encoder.layer.2.attention.self.key	&	18.0    \\
bert.encoder.layer.2.attention.self.value	&	15.0    \\
bert.encoder.layer.2.attention.output.dense	&	16.0    \\
bert.encoder.layer.2.intermediate.dense	&	17.0    \\
bert.encoder.layer.2.output.dense	&	20.0    \\
bert.encoder.layer.3.attention.self.query	&	18.0    \\
bert.encoder.layer.3.attention.self.key	&	18.0    \\
bert.encoder.layer.3.attention.self.value	&	16.0    \\
bert.encoder.layer.3.attention.output.dense	&	16.0    \\
bert.encoder.layer.3.intermediate.dense	&	17.0    \\
bert.encoder.layer.3.output.dense	&	20.0    \\
bert.encoder.layer.4.attention.self.query	&	17.0    \\
bert.encoder.layer.4.attention.self.key	&	17.0    \\
bert.encoder.layer.4.attention.self.value	&	16.0    \\
bert.encoder.layer.4.attention.output.dense	&	17.0    \\
bert.encoder.layer.4.intermediate.dense	&	17.0    \\
bert.encoder.layer.4.output.dense	&	20.0    \\
bert.encoder.layer.5.attention.self.query	&	17.0    \\
bert.encoder.layer.5.attention.self.key	&	17.0    \\
bert.encoder.layer.5.attention.self.value	&	15.0    \\
bert.encoder.layer.5.attention.output.dense	&	16.0    \\
bert.encoder.layer.5.intermediate.dense	&	17.0    \\
bert.encoder.layer.5.output.dense	&	20.0    \\
bert.encoder.layer.6.attention.self.query	&	17.0    \\
bert.encoder.layer.6.attention.self.key	&	17.0    \\
bert.encoder.layer.6.attention.self.value	&	15.0    \\
bert.encoder.layer.6.attention.output.dense	&	16.0    \\
bert.encoder.layer.6.intermediate.dense	&	17.0    \\
bert.encoder.layer.6.output.dense	&	20.0    \\
bert.encoder.layer.7.attention.self.query	&	17.0    \\
bert.encoder.layer.7.attention.self.key	&	17.0    \\
bert.encoder.layer.7.attention.self.value	&	15.0    \\
bert.encoder.layer.7.attention.output.dense	&	16.0    \\
bert.encoder.layer.7.intermediate.dense	&	17.0    \\
bert.encoder.layer.7.output.dense	&	19.0    \\
bert.encoder.layer.8.attention.self.query	&	17.0    \\
bert.encoder.layer.8.attention.self.key	&	17.0    \\
bert.encoder.layer.8.attention.self.value	&	16.0    \\
bert.encoder.layer.8.attention.output.dense	&	16.0    \\
bert.encoder.layer.8.intermediate.dense	&	18.0    \\
bert.encoder.layer.8.output.dense	&	20.0    \\
bert.encoder.layer.9.attention.self.query	&	16.0    \\
bert.encoder.layer.9.attention.self.key	&	17.0    \\
bert.encoder.layer.9.attention.self.value	&	15.0    \\
bert.encoder.layer.9.attention.output.dense	&	17.0    \\
bert.encoder.layer.9.intermediate.dense	&	18.0    \\
bert.encoder.layer.9.output.dense	&	20.0    \\
bert.encoder.layer.10.attention.self.query	&	16.0    \\
bert.encoder.layer.10.attention.self.key	&	16.0    \\
bert.encoder.layer.10.attention.self.value	&	16.0    \\
bert.encoder.layer.10.attention.output.dense	&	17.0    \\
bert.encoder.layer.10.intermediate.dense	&	18.0    \\
bert.encoder.layer.10.output.dense	&	20.0    \\
bert.encoder.layer.11.attention.self.query	&	17.0    \\
bert.encoder.layer.11.attention.self.key	&	17.0    \\
bert.encoder.layer.11.attention.self.value	&	16.0    \\
bert.encoder.layer.11.attention.output.dense	&	17.0    \\
bert.encoder.layer.11.intermediate.dense	&	18.0    \\
bert.encoder.layer.11.output.dense	&	20.0    \\
bert.pooler.dense	&	15.0    \\
classifier	&	8.0 \\
\label{tab:bert_mrpc_bw_last_step}\\
\end{supertabular}
\end{center}

\subsection{BERT on SQUADv1.1}

\begin{figure}[hbt]
\centering
\includegraphics[width=0.6\linewidth]{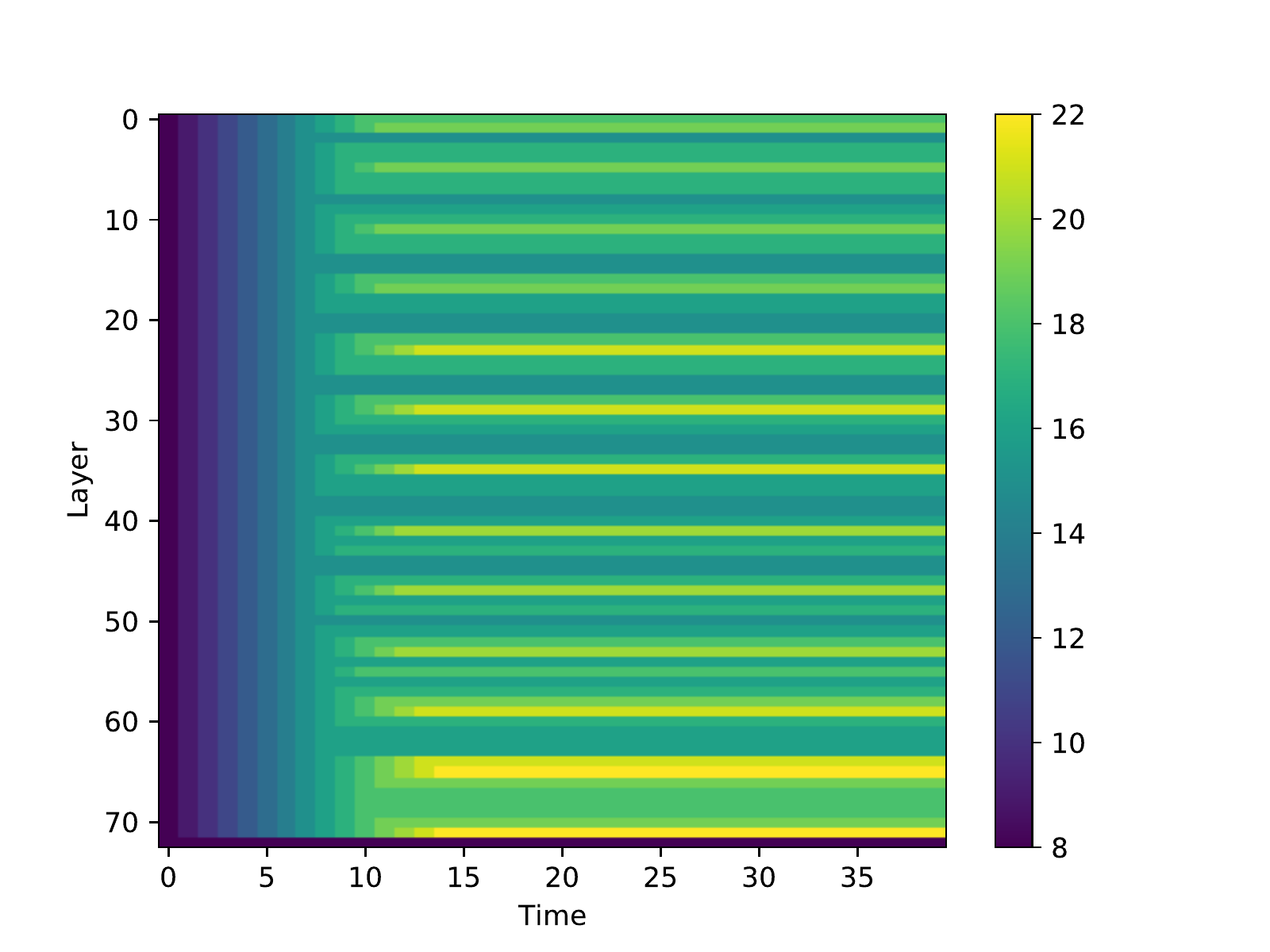}
\caption{BERT on SQaADv1.1, history of bitwidth. x-axis stands for time, y-axis stands for different layers. First layer is at the top. Each time tick represents 369 steps}
\label{fig:bert_squad_bw_history}
\end{figure}

\begin{center}
\tablefirsthead{
\multicolumn{1}{l}{Layer name} & \multicolumn{1}{l}{Bitwidth} \\ \hline
}
\tablehead{
\multicolumn{2}{l}{\small\sl continued from previous page}\\ \hline
\multicolumn{1}{l}{Layer name} & \multicolumn{1}{l}{Bitwidth} \\ \hline
}
\tabletail{%
\hline
\multicolumn{2}{r}{\small\sl continued on next page}\\
}
\tablelasttail{\hline}
\tablecaption{BERT on SQUADv1.1, bitwidth at last step}
\begin{supertabular}{l l}
bert.encoder.layer.0.attention.self.query	&	18.0	\\
bert.encoder.layer.0.attention.self.key	&	19.0	\\
bert.encoder.layer.0.attention.self.value	&	15.0	\\
bert.encoder.layer.0.attention.output.dense	&	17.0	\\
bert.encoder.layer.0.intermediate.dense	&	17.0	\\
bert.encoder.layer.0.output.dense	&	19.0	\\
bert.encoder.layer.1.attention.self.query	&	17.0	\\
bert.encoder.layer.1.attention.self.key	&	17.0	\\
bert.encoder.layer.1.attention.self.value	&	15.0	\\
bert.encoder.layer.1.attention.output.dense	&	16.0	\\
bert.encoder.layer.1.intermediate.dense	&	17.0	\\
bert.encoder.layer.1.output.dense	&	19.0	\\
bert.encoder.layer.2.attention.self.query	&	17.0	\\
bert.encoder.layer.2.attention.self.key	&	17.0	\\
bert.encoder.layer.2.attention.self.value	&	15.0	\\
bert.encoder.layer.2.attention.output.dense	&	15.0	\\
bert.encoder.layer.2.intermediate.dense	&	18.0	\\
bert.encoder.layer.2.output.dense	&	19.0	\\
bert.encoder.layer.3.attention.self.query	&	16.0	\\
bert.encoder.layer.3.attention.self.key	&	16.0	\\
bert.encoder.layer.3.attention.self.value	&	15.0	\\
bert.encoder.layer.3.attention.output.dense	&	15.0	\\
bert.encoder.layer.3.intermediate.dense	&	18.0	\\
bert.encoder.layer.3.output.dense	&	21.0	\\
bert.encoder.layer.4.attention.self.query	&	17.0	\\
bert.encoder.layer.4.attention.self.key	&	17.0	\\
bert.encoder.layer.4.attention.self.value	&	15.0	\\
bert.encoder.layer.4.attention.output.dense	&	15.0	\\
bert.encoder.layer.4.intermediate.dense	&	18.0	\\
bert.encoder.layer.4.output.dense	&	21.0	\\
bert.encoder.layer.5.attention.self.query	&	17.0	\\
bert.encoder.layer.5.attention.self.key	&	16.0	\\
bert.encoder.layer.5.attention.self.value	&	15.0	\\
bert.encoder.layer.5.attention.output.dense	&	15.0	\\
bert.encoder.layer.5.intermediate.dense	&	17.0	\\
bert.encoder.layer.5.output.dense	&	21.0	\\
bert.encoder.layer.6.attention.self.query	&	16.0	\\
bert.encoder.layer.6.attention.self.key	&	16.0	\\
bert.encoder.layer.6.attention.self.value	&	15.0	\\
bert.encoder.layer.6.attention.output.dense	&	15.0	\\
bert.encoder.layer.6.intermediate.dense	&	16.0	\\
bert.encoder.layer.6.output.dense	&	20.0	\\
bert.encoder.layer.7.attention.self.query	&	16.0	\\
bert.encoder.layer.7.attention.self.key	&	17.0	\\
bert.encoder.layer.7.attention.self.value	&	15.0	\\
bert.encoder.layer.7.attention.output.dense	&	15.0	\\
bert.encoder.layer.7.intermediate.dense	&	17.0	\\
bert.encoder.layer.7.output.dense	&	20.0	\\
bert.encoder.layer.8.attention.self.query	&	16.0	\\
bert.encoder.layer.8.attention.self.key	&	17.0	\\
bert.encoder.layer.8.attention.self.value	&	15.0	\\
bert.encoder.layer.8.attention.output.dense	&	16.0	\\
bert.encoder.layer.8.intermediate.dense	&	18.0	\\
bert.encoder.layer.8.output.dense	&	20.0	\\
bert.encoder.layer.9.attention.self.query	&	16.0	\\
bert.encoder.layer.9.attention.self.key	&	18.0	\\
bert.encoder.layer.9.attention.self.value	&	16.0	\\
bert.encoder.layer.9.attention.output.dense	&	17.0	\\
bert.encoder.layer.9.intermediate.dense	&	19.0	\\
bert.encoder.layer.9.output.dense	&	21.0	\\
bert.encoder.layer.10.attention.self.query	&	17.0	\\
bert.encoder.layer.10.attention.self.key	&	16.0	\\
bert.encoder.layer.10.attention.self.value	&	16.0	\\
bert.encoder.layer.10.attention.output.dense	&	16.0	\\
bert.encoder.layer.10.intermediate.dense	&	21.0	\\
bert.encoder.layer.10.output.dense	&	22.0	\\
bert.encoder.layer.11.attention.self.query	&	19.0	\\
bert.encoder.layer.11.attention.self.key	&	18.0	\\
bert.encoder.layer.11.attention.self.value	&	18.0	\\
bert.encoder.layer.11.attention.output.dense	&	18.0	\\
bert.encoder.layer.11.intermediate.dense	&	19.0	\\
bert.encoder.layer.11.output.dense	&	22.0	\\
qa\_outputs	&	8.0	\\
\label{tab:bert_squad_bw_last_step}\\
\end{supertabular}
\end{center}

\end{document}